\PassOptionsToPackage{table}{xcolor}
\documentclass[runningheads]{llncs}

\usepackage[preprint]{colm} 

\setcitestyle{authoryear}

\usepackage{graphicx}
\usepackage{booktabs}
\usepackage[accsupp]{axessibility}
\usepackage{xltabular} % The missing piece
\usepackage{microtype}
\usepackage{tabularx}
\usepackage{amssymb}
\usepackage{paralist}
\usepackage{todonotes}
\usepackage{breakurl}
\usepackage{subcaption}
\usepackage{latexsym}
\usepackage{amsmath}
\usepackage{float}
\usepackage{footnote}
\usepackage{enumitem}
\usepackage{bm}
\usepackage{arydshln}
\usepackage{multicol}
\usepackage{multirow}
\usepackage{colortbl}
\usepackage{bbding}
\usepackage{makecell}
\usepackage{mathtools}
\usepackage{imakeidx}
\usepackage{longtable}
\usepackage{tabularx}
\usepackage{wrapfig}
\usepackage{rotating}
\usepackage[edges]{forest}
\usepackage[normalem]{ulem}
\usepackage{CJKutf8}
\usepackage{awesomebox}
\usepackage[most]{tcolorbox}
\usepackage{graphicx}
\usepackage{booktabs}
\usepackage[table]{xcolor}
\usepackage{adjustbox}

\RequirePackage{xspace}
\makeatletter
\DeclareRobustCommand\onedot{\futurelet\@let@token\@onedot}
\def\@onedot{\ifx\@let@token.\else.\null\fi\xspace}

\makeatother

\usepackage{pifont}
\usepackage{lscape}
\usepackage{algorithm}
\usepackage{algpseudocode}

\usepackage{fancyhdr}
\renewcommand{\headrulewidth}{1pt}
\usepackage{fancyvrb}
\usepackage{fvextra}
\usepackage{amsfonts}
\usepackage{wrapfig}

\usepackage{hyperref}
\usepackage{orcidlink}

\definecolor{mydarkblue}{rgb}{0,0.08,0.45}
\definecolor{wkblue}{rgb}{0.2, 0.3, 0.6}
\definecolor{meta-color}{rgb}{0.5, 0.5, 0.5}
\definecolor{darkblue}{rgb}{0, 0, 0.5}
\definecolor{geovistagray}{gray}{0.95}
\definecolor{myblue}{rgb}{0.9, 0.1, 0.94}
\definecolor{mygreen}{rgb}{0.64, 0.56, 0.88}
\definecolor{myyellow}{rgb}{0.68, 0.6, 0.1}
\definecolor{fancygreen}{rgb}{0.33, 0.68, 0.20}
\definecolor{salmon}{rgb}{0.94, 0.52, 0.49}
\definecolor{tablegreen}{rgb}{0.82, 0.94, 0.75}
\definecolor{tableblue}{rgb}{0.81, 0.90, 0.94}
\definecolor{tablered}{rgb}{0.97, 0.85, 0.85}
\definecolor{tableorange}{rgb}{0.96, 0.85, 0.81}
\definecolor{bestcolor}{RGB}{210, 222, 239}
\definecolor{secondcolor}{RGB}{234, 239, 247}
\definecolor{thirdcolor}{RGB}{193, 214, 229}
\definecolor{line-blue}{RGB}{243, 248, 252}
\definecolor{line-green}{RGB}{200,242,200}
\definecolor{line-red}{RGB}{255,215,215}
\definecolor{line-gray}{RGB}{242, 242, 242}
\definecolor{sensepurple}{HTML}{5D2DD6}

\newenvironment{itemize*}%
 {\leftmargini=10pt\begin{itemize}%
  \setlength{\itemsep}{0pt}%
  \setlength{\parskip}{0pt}%
  }%
 {\end{itemize}}
\newenvironment{enumerate*}%
 {\begin{enumerate}%
  \setlength{\itemsep}{0pt}%
  \setlength{\parskip}{0pt}}%
 {\end{enumerate}}

\begin{document}

\title{GEM: Generative Supervision Helps Embodied Intelligence}

\titlerunning{GEM}

\author{\textbf{Ruowen Zhao$^{1}$, Bangguo Li$^{1}$, Zuyan Liu$^{1,2,\dagger}$,}
\textbf{Yinan Liang$^{1}$, Junliang Ye$^{1}$, Fangfu Liu$^{1}$,} \\\textbf{Diankun Wu$^{1}$, Zhengyi Wang$^{1}$,}
\textbf{Xumin Yu$^{2}$, Yongming Rao$^{2,\ddagger}$, Han Hu$^{2}$, Jun Zhu$^{1,\ddagger}$}\\[0.5em]
$^1$ Tsinghua University \quad $^2$ Tencent Hunyuan}

\authorrunning{Zhao et al.}

\maketitle

\makeatletter
\def\@makefnmark{}
\makeatother
\footnotetext{$^{\dagger}$ Project Lead. $^{\ddagger}$ Corresponding author.}

% \vspace{-0.5em}

\begin{abstract}
Embodied Vision-Language Models (VLMs) have demonstrated impressive performance and generalization in robotics, particularly within Vision-Language-Action frameworks.   However, a significant gap remains between the high-level semantic focus of standard text-guided pre-training paradigms and the low-level spatial and physical knowledge critical for execution in embodied environments.   In this paper, we introduce \textbf{GEM}, a \textbf{G}enerative-supervised \textbf{Em}bodied vision-language Model designed to bridge this divide.   We propose integrating a depth map generation task directly into the VLM pre-training phase.   By training this generative objective jointly with the main model, we observe substantial improvements in embodied intelligence, significantly enhancing both semantic understanding and physical operation capabilities.   To support this paradigm, we curate and release GEM-4M, a comprehensive large-scale dataset featuring a mixture of grounding, reasoning, and planning data paired with high-quality depth supervision.   Extensive experiments demonstrate that GEM achieves state-of-the-art results across diverse embodied benchmarks.   Furthermore, our deployed action model, GEM-VLA, exhibits vastly superior task execution abilities in both simulation environments and real-world evaluations. Code, models, and datasets are available at \url{https://zhaorw02.github.io/GEM/}.
\keywords{Embodied Intelligence \and Vision-Language Models  \and Vision-Language-Action Models}
\end{abstract}

\section{Introduction}

Recent advancements in Vision-Language Models (VLMs)~\citep{bai2025qwen3,wang2025internvl3,li2024llava,beyer2024paligemma,liu2023visual} have unlocked remarkable capabilities in embodied understanding, encompassing critical skills such as spatial recognition, physical grounding, and complex task planning.   By effectively aligning visual perception with natural language reasoning, these models~\citep{yang2025vlaser,hao2025mimo,ji2025robobrain,liu2025towards} have emerged as robust foundation architectures for Vision-Language-Action (VLA) frameworks~\citep{kim2024openvla,intelligence2025pi_,team2024octo,brohan2022rt}. Consequently, Embodied VLMs are increasingly being leveraged to drive a massive array of downstream operational tasks, demonstrating potential for generalization and autonomous execution within dynamic, real-world physical environments.

\begin{center}
    \centering
    \vspace{-0.5cm}
    \includegraphics[width=1.0\linewidth]{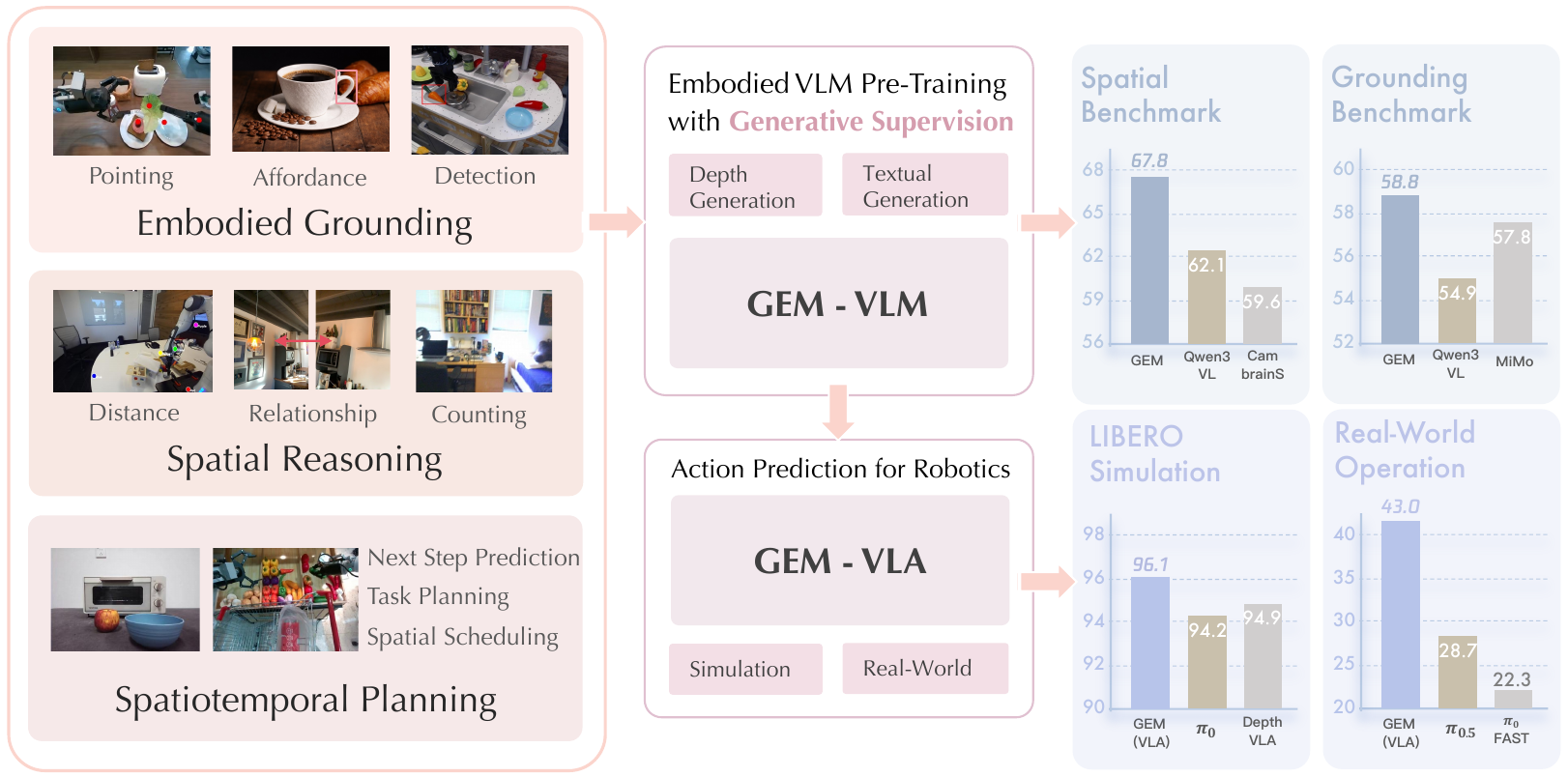}%
    \captionof{figure}
    {
\textbf{Overview of GEM.} GEM is a generative-supervised embodied VLM that strengthens semantic reasoning and physical grounding by combining language modeling with an auxiliary depth-generation objective (center). Trained on the high-quality, large-scale pre-training datasets spanning diverse embodied tasks (left), GEM achieves strong performance across a wide range of embodied benchmarks. Based on the architecture of GEM, the extending GEM-VLA attains state-of-the-art success rates on LIBERO and generalizes well to real-world robot manipulation (right).
    }
    \label{fig:teaser}
\end{center}%

Despite these foundational successes, the predominant paradigm for training embodied VLMs~\citep{ji2025robobrain,yang2025vlaser,dang2026rynnbrain,hao2025mimo,azzolini2025cosmos} relies heavily on scaling up massive visual question answering datasets~\citep{sermanet2024robovqa,yuan2024robopoint,yang2025vlaser,qu2025eo,chen2025robo2vlm}.  While this approach effectively boosts performance on high-level semantic benchmarks and passive comprehension tasks~\citep{zhou2025roborefer,yuan2024robopoint,song2025robospatial,tong2024cambrian,yang2024think}, it inherently creates a disconnect from the physical constraints of real-world applications.  Because these datasets primarily emphasize descriptive reasoning over active, physical interaction, a critical bottleneck emerges: superior semantic comprehension does not invariably translate to proficient task execution in complex, real-world environments.  Conversely, alternative lines of research~\citep{qu2025spatialvla,yuan2025depthvla,li2026pointvla,zheng2024tracevla} attempt to bridge this gap by explicitly integrating spatial, temporal, and low-level physical knowledge directly into downstream VLA models~\citep{kim2024openvla,team2024octo} to enhance operational performance.  However, these low-level physical priors are typically injected late in the pipeline or treated as separate entities from the broad textual pre-training data.  This isolates critical physical grounding from the rich, open-vocabulary semantic guidance of linguistic models, preventing the development of a truly unified, embodied representation.  Consequently, a critical and emerging question arises: how can we seamlessly embed essential spatial and physical knowledge directly into the foundational pre-training phase of vision-language models, such that it tangibly elevates both abstract semantic reasoning and actionable, real-world operational intelligence?

To overcome these limitations, we propose \textbf{GEM}, a \textbf{G}enerative-supervised \textbf{Em}bodied vision-language model. To effectively capture fine-grained structural details and complete spatial and geometric relations within visual scenes, we establish depth map prediction~\citep{lin2025depth} as an intrinsic generative target.  This is achieved through a novel hybrid autoregressive-diffusion architecture~\citep{chen2025blip3o,wu2025omnigen2} designed to seamlessly blend generative and representational supervision.  Specifically, our approach conditions a diffusion transformer~\citep{peebles2023scalable,lipman2022flow} on the hidden visual features extracted by an auto-regressive understanding model~\citep{bai2025qwen3} to synthesize accurate depth maps.  To facilitate this integration, we implement a progressive training strategy that initially stabilizes the generation module before jointly optimizing for both depth synthesis and linguistic knowledge acquisition.  Furthermore, to synergize with our architectural and training advancements, we introduce GEM-4M, a high-quality, large-scale embodied pre-training dataset.  GEM-4M encompasses extensive embodied question-answering pairs that rigorously cover physical grounding, spatial-temporal planning, and physical reasoning tasks.  Ultimately, the comprehensive spatial and semantic representations learned by the GEM architecture can be effortlessly extended into a VLA model, denoted as GEM-VLA, facilitating robust, autonomous performance in real-world robotic deployments.

Extensive experimental evaluations demonstrate that GEM and GEM-VLA show remarkable performance under a wide range of benchmarks from recognition to real-world operations. GEM establishes a new state-of-the-art, consistently outperforming leading open-source general-purpose models, as well as spatial and embodied specialists, on key reasoning benchmarks. Specifically, GEM attains the highest overall scores on the challenging spatial-related benchmarks~\citep{yang2024think,yang2025mmsi,du2024embspatial,tong2024cambrian} and shows large gains over its initialization backbones.  For instance, the VSI-Bench~\citep{yang2024think} score improves from 50.4 to 62.8 for the 2B model and from 57.9 to 70.6 for the 8B model. On benchmarks that require fine-grained spatial grounding~\citep{zhou2025roborefer,yuan2024robopoint,song2025robospatial}, GEM far exceeds the performance of the strong proprietary baseline, Gemini-3-Pro, by 10\%. Furthermore, our vision-language-action model, GEM-VLA, achieves a record-breaking 96.1\% average success rate on the LIBERO~\citep{liu2023libero} benchmark, outperforming standard VLAs such as $\pi_0$ and spatial-enhanced VLAs~\citep{qu2025spatialvla,yuan2025depthvla}. GEM-VLA also transfers robustly to challenging real-world settings and surpasses recent methods~\citep{pertsch2025fast,intelligence2025pi0.5_} with an average success rate of 43\%, marking a substantial improvement over the previous state-of-the-art's 28.7\%.

\section{Related Work}

\nocite{zhao2025deepmesh,zhao2024flexidreamer,ye2025shapellm,tencent2026hyembodied05}
\subsection{Vision-Language Models for Embodied Intelligence}

Enhancing the embodied reasoning capabilities of state-of-the-art Vision-Language Models (VLMs) has become a central research focus. A number of data-driven methodologies have emerged to support such reasoning capabilities, including object affordances for manipulation, object counting, spatial relationship understanding, and action planning that determines subsequent steps based on the current states. For instance, some studies~\citep{team2025gemini,azzolini2025cosmos,luo2025visual,molmoact2025,qu2025eo,yang2025vlaser,hao2025mimo,eo1} contribute curated datasets specifically tailored for embodied tasks, emphasizing multi-modal understanding and action-aware visual-language alignment. Additionally, other works~\citep{ji2025robobrain,yuan2025embodied,dang2026rynnbrain,zhou2025roborefer,zhang2025pelican} construct synthetic spatiotemporal reasoning datasets enriched with Chain-of-Thought (CoT) annotations~\citep{wei2022chain} and then incorporate Reinforcement Fine-Tuning (RFT)~\citep{shao2024deepseekmath} to further refine reasoning performance of Embodied VLMs. Nevertheless, existing approaches mainly focus on high-level semantic understanding, while overlooking the explicit modeling of fine-grained structural information in visual inputs. As a result, the visual features fail to preserve fine-grained geometric cues, leading to ambiguous spatial relationships. This issue is particularly critical for embodied tasks, where precise perception of object geometry and relative distances is essential for robust manipulation and interaction. In this paper, we imitate this issue by introducing generative supervision to facilitate the fusion of structural and semantic features for more comprehensive embodied reasoning.

\subsection{Spatial-Aware Vision-Language-Action Models}

Robotic manipulation has evolved from single-task specialists to generalist models trained on broad, diverse datasets. Fueled by advances in VLMs~\citep{beyer2024paligemma,bai2025qwen3,wang2025internvl3,comanici2025gemini}, and large-scale robot action datasets~\citep{bu2025agibot_iros,o2024open,wu2024robomind,khazatsky2024droid,wu2025robocoin}, this evolution has given rise to the architecture of Vision-Language-Action (VLA) models~\citep{brohan2022rt,kim2024openvla,team2024octo,intelligence2025pi_,cheang2025gr,li2023vision,liu2026rdt2,wen2025tinyvla,liu2025hybridvla}, which integrate the VLM backbone with robot action output head. Inheriting the rich perceptual and linguistic representations of pretrained VLMs, VLA models demonstrate improved adaptability and zero-shot capabilities in interpreting and executing human instructions. Despite their promising performance, current VLAs are primarily confined to 2D observation inputs and lack precise perception and comprehension of the 3D physical world. To bridge this gap, early efforts augmented VLAs with 3D or 2.5D inputs ~\citep{li2026pointvla,ze20243d,zhen20243d,li20253ds,zheng2024tracevla}. However, such approaches suffer from expensive computational and data acquisition costs. More recent works~\citep{li2025spatial,qu2025spatialvla,yuan2025depthvla,wu2026pragmatic,song2025reconvla} instead explore various implicit enhancement strategies that implicit enhancement strategies that integrate global spatial context into the semantic representations from 2D observations, to inject geometric priors. Nevertheless, these methods mainly rely on simple feature fusion, which limits their ability to substantially improve spatial perception. Other works~\citep{dreamvla25,zhao2025cot,zhang2025up,jiang2025rynnvla,cen2025worldvla,wang2025unified,hu2024video,liao2025genie,lv2025f1} incorporate generative world models that predict future frames or states to inject world knowledge. Although this improves planning by simulating futures, it contributes little to strengthening the geometric encoding of the current scene. Overall, enhancing VLAs with robust and physically grounded perception of the real world remains an open and challenging problem.

\section{Method}

In this section, we detail our design of GEM's overall framework. We elaborate our architecture design in Sec. \ref{sec:arch} and progressive training pipeline in Sec.\ref{sec:train}. Then we describe the construction of our training dataset GEM-4M in Sec.\ref{sec:dataset}. Finally, we explain how we extend our model to a VLA framework for downstream robot tasks in Sec. \ref{sec:vla}.

\subsection{Architecture} \label{sec:arch}

 % Since existing VLMs are pre-trained solely on RGB images, they focus primarily on semantic features for inference, lacking the structural information needed for accurate 3D perception. This limits their performance in spatial reasoning, especially for embodied tasks. Recent methods~\citep{wang2023chat,hong20233d,fu2024scene} have incorporated explicit 3D modalities and expert encoders as additional inputs to enhance global spatial awareness, but their high computational costs limit practical applicability. To address this, we propose depth generative supervision, where visual features are used as condition to generate depth images, enabling the fusion of both semantic and structural information within the visual representation space.

In current VLMs, given an instruction $l$ and visual input $o$, the VLM backbone $M_\theta$ encodes them into multimodal token representations $\mathbf{h} = (\mathbf{h}_o, \mathbf{h}_l) = M_{\theta}(o, l)$ at its final layer. Then they are trained to maximize the likelihood of the target token sequence $y$, typically using a cross-entropy objective for supervised fine-tuning:
 \begin{equation}
     \mathcal{L}_{\text{CE}} = -\sum_{i=1}^{T} \log p_\theta(y_i | y_{<i}, \mathbf{h}_o, \mathbf{h}_l)
 \end{equation}
 
 This objective helps the models align visual token features with text and perform semantic understanding tasks. Despite demonstrating outstanding performance in various visual tasks, their spatial reasoning ability, particularly in embodied scenarios, is limited because $\mathbf{h}_o$ contains only semantic information from $o$ and lacks sufficient physical structural cues for accurate spatial understanding and manipulation in real-world environments.
 
\begin{figure}[tb]
  \centering
  \includegraphics[width=\linewidth]{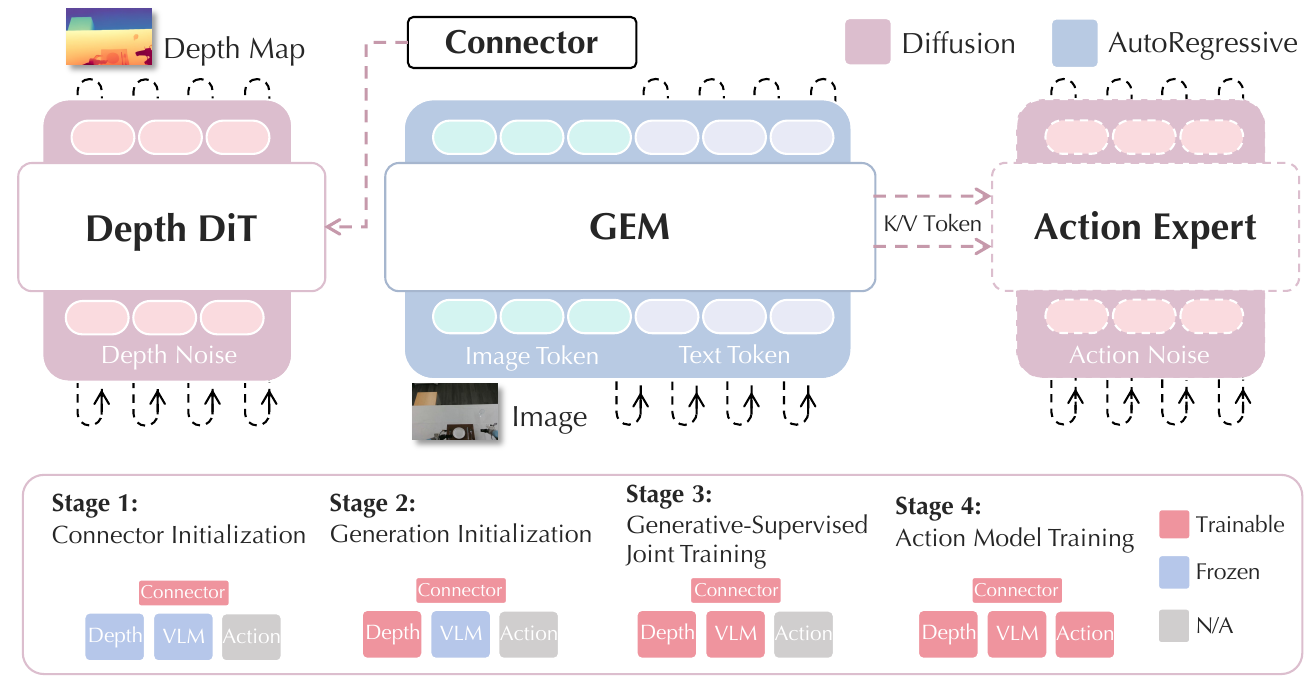}
  \caption{\textbf{Architecture of GEM.} GEM augments a VLM backbone with a DiT-based depth generator conditioned on the backbone’s final-layer visual tokens. We adopt a progressive training paradigm: (i) initialize the connector, (ii) warm up the depth generator, (iii) perform end-to-end joint training, and (iv) train an autoregressive action expert on GEM’s multimodal tokens. Building on GEM, the GEM-based VLA predicts continuous actions from these representations, improving robot manipulation. }
  \label{fig:pipeline}
\end{figure}

 To address this, we introduce a depth generative objective for supervision. As illustrated in Figure \ref{fig:pipeline}, GEM consists of a VLM backbone $M_\theta$, a lightweight connector $C_{\phi}$ and a Diffusion Transformer (DiT)-based depth generative head $G_{\psi}$. In our design, the visual tokens in $\mathbf{h}$, denoted as $\mathbf{h}_o$, are projected into a conditional embedding space via the connector: $\mathbf{c}=C_{\phi}(\mathbf{h}_o)$. We propose to utilize $\mathbf{c}$ as the condition for the generative head to reconstruct the observation $o$'s depth map $d$. We then employ a flow matching objective $\mathcal{L}_\textnormal{flow}$ to optimize the generative head, which learns the vector field $v_t$ at each timestep $t \sim \mathcal{U}(0,1)$ that transforms a noised distribution $\mathbf{x}_t$ into the ground-truth depth $d$:
 \begin{equation}
      \mathcal{L}_\textnormal{flow} = \mathbb{E}_{d, t \sim \mathcal{U}(0,1), \epsilon \sim \mathcal{N}(\mathbf{0}, \mathbf{I})}\left[\|\mathbf{v}_t(\mathbf{x}_t, \mathbf{c}) - \mathbf{u}_t(\mathbf{x}_t \mid d)\|^2\right]
\end{equation}
where $\mathbf{u}_t(\mathbf{x}_t \mid d)$ is the ground-truth velocity field that transforms $\mathbf{x}_t$ into depth $d$. We combine this generative supervision loss with $\mathcal{L}_\textnormal{CE}$ to allow $\mathbf{h}_o$ to encode adequate structural information for depth generation, as well as sufficient semantic information for inference.

\subsection{Progressive Training Recipe} \label{sec:train}
Since there is a gap between the backbone’s output space and the DiT’s input space, directly training the overall framework end-to-end may cause modality interference between the generative head and the VLM backbone, leading to unstable convergence. To address this, we adopt a progressive training recipe to bridge the gap between the two feature spaces effectively. Specifically, the training pipeline is divided into the following three distinct phases:

\subsubsection{Stage 1: Connector Initialization}

In the first stage, we freeze both the pre-trained VLM backbone and the DiT generative head, and only optimize the connector for preliminary feature alignment. The connector projects the backbone's semantic representations into the DiT’s input feature space to establish a stable start for later training stages. At this stage, only the generative objective $\mathcal{L}_\textnormal{flow}$ is used.

\subsubsection{Stage 2: Generative Head Initialization}
After preliminary feature alignment, the generative head has not yet adapted to the conditioning features from the VLM backbone. Therefore, we freeze the backbone and only optimize both the connector and DiT head to equip the depth generative head with basic image generation ability. At this stage, the generative objective $\mathcal{L}_\textnormal{flow}$ is used solely to transform high-level semantic features into fine-grained structure features, building the foundation for subsequent joint training.

\subsubsection{Stage 3: Generative-Supervised Joint Training}
In the final stage, we perform end-to-end generative-supervised joint training. Since the first two stages have established a stable initialization, we unfreeze the trainable parameters of the entire framework, including VLM backbone, connector, and DiT head, to foster synergy between the backbone's semantic understanding and DiT's generative capability. This allows VLM not only to understand semantics but also to refine its representations to be more structure-aware, capturing subtle geometric cues and spatial relationships. At this stage, both cross-entropy text loss $\mathcal{L}_\textnormal{CE}$ and flow-matching generative loss $\mathcal{L}_\textnormal{flow}$ supervise the training process, with the total loss defined as $ \mathcal{L}_{\textnormal{total}} = \mathcal{L}_{\textnormal{CE}} + \lambda \mathcal{L}_{\textnormal{flow}}$, where $\lambda$ is the balancing weight.

\subsection{Dataset} \label{sec:dataset}

To advance the capability of GEM in perception and reasoning real-world scenarios grounded in physical knowledge, we construct a high-quality, large-scale question-answer (QA) dataset, GEM-4M, for supervised fine-tuning. Here we present an overview of the data building engine and sources, while more details about the construction methodologies are provided in the supplementary materials.

\subsubsection{Embodied Grounding Data}
To enhance the model's object recognition and localization capacities in embodied scenarios, we collect 1M high-quality question-answer pairs to support multiple grounding tasks, including open-vocabulary object detection with bounding boxes, localizing objects from instructions, and recognizing object affordances. These data are sourced from several publicly available embodied grounding datasets, such as PACO-LVIS~\citep{ramanathan2023paco}, RoboPoint~\citep{yuan2024robopoint}, RoboAfford~\citep{hao2025roboafford++}, ShareRobot~\citep{ji2025robobrain}, and Roborefit~\citep{lu2023vl}. Additionally, to ensure grounding in physical manipulation scenarios, we generate approximately 100k point and bounding box annotations from open-source robot action datasets~\citep{wu2024robomind, khazatsky2024droid, bu2025agibot_iros, o2024open} using SAM3~\citep{carion2025sam3segmentconcepts}. This combination of open-source and self-curated data covers a wide range of scenarios, enhancing the diversity and generalization of visual grounding in real-world embodied environments. To handle varying image resolutions, both bounding boxes and points are normalized to the range $[0, 1000]$ to ensure consistency.

\subsubsection{Physical, Spatial Reasoning Data}

This category of data aims to help the model build a foundational understanding of the physical world, such as measurement estimation and spatiotemporal reasoning. Specifically, We incorporate open-source spatial datasets, including MindCube~\citep{yin2025spatial}, ViCA~\citep{feng2025visuospatial}, SPAR~\citep{zhang2025flatland}, and VSI-590K~\citep{yang2025cambrians}, to support 3D spatial reasoning and physical attribute perception. Additionally, we also augment these datasets with 100k manually annotated spatial understanding samples from publicly available 3D scene datasets~\citep{dai2017scannet,yeshwanth2023scannet++,baruch2021arkitscenes}, following the data processing pipeline proposed in VSI-Bench~\citep{yang2025cambrians}. To improve spatiotemporal abilities especially in robot tasks, we integrate 1 million question-answer pairs aggregated from multiple publicly available datasets, such as RoboVQA~\citep{sermanet2024robovqa}, Robo2VLM~\citep{chen2025robo2vlm}, and RefSpatial~\citep{zhou2025roborefer}. The integration of these diverse, high-quality data sources strengthens the model's spatial awareness and boosts performance in complex embodied reasoning tasks. 

\subsubsection{Spatiotemporal Planning Data} 

% Spatial-Temporal

To equip the embodied brain with the ability to plan sub-tasks and forecast the trajectory of each atomic action, we collect data from public robot datasets~\citep{wu2025robocoin, bu2025agibot_iros, wu2024robomind} with sub-task annotations and construct question-answer pairs. We extract individual frames from entire egocentric videos based on sub-task annotations and identify the manipulated object in each sub-task description using Qwen3~\citep{yang2025qwen3}. We then use SAM3~\citep{carion2025sam3segmentconcepts} to generate object masks and track their trajectory using CoTracker3~\citep{karaev2025cotracker3}. Finally, based on the sub-task descriptions and visualized trajectories, we create sub-task and trajectory planning question-answer pairs respectively following the RoboVQA~\citep{sermanet2024robovqa} and MolmoACT~\citep{molmoact2025} templates, resulting in a dataset of approximately 50K samples. The integration of these spatiotemporal data allows the model to combine basic skills, generalize to new scenarios, and plan actions effectively.

\subsection{Expanding to Vision-Language Action Model} \label{sec:vla}

We integrate GEM into a VLA framework to evaluate its transfer to robotic manipulation. As illustrated in Figure~\ref{fig:pipeline}, we integrate a Diffusion Transformer (DiT)-based action expert, denoted as $A_{\omega}$, to generate continuous actions from multi-modal observations via a diffusion policy. We extract the key–value tokens of the multimodal observation history $\mathcal{O}$ from the attention blocks in backbone $M_{\theta}$ and use it as the conditioning representation $\mathbf{c}_{\textnormal{act}}$ for the action expert $A_{\omega}$, to bridge high-level reasoning capabilities and low-level action generation. We perform end-to-end joint optimization of the VLM $M_{\theta}$, the depth generative head $G_{\psi}$, and the action expert $A_{\omega}$ using a combination of both depth and action generative objectives. Specifically, the action objective $\mathcal{L}_{\text{action}}$ aims to predict the vector field $\mathbf{v}_t$ at each timestep $t \sim \mathcal{U}(0,1)$ that transforms a noisy action state $\mathbf{a}_t = (1 - t)\epsilon + t\mathbf{a}$ into the ground-truth action chunk $\mathbf{a}$:
\begin{equation}
    \mathcal{L}_{\text{action}} = \mathbb{E}_{\mathcal{O}, \mathbf{a}, \epsilon \sim \mathcal{N}(\mathbf{0}, \mathbf{I}), t \sim \mathcal{U}(0,1)} \left[ \lVert \mathbf{v}_{t}({\mathbf{a}}_t, \mathbf{c}_{\textnormal{act}}) - \mathbf{u}_t(\mathbf{a}_t \mid \mathbf{a}) \rVert_2^2 \right]
    \label{eq:action}
\end{equation}
The total loss is then defined as: $ \mathcal{L}_{\textnormal{total}} = \mathcal{L}_{\textnormal{action}} + \lambda \mathcal{L}_{\textnormal{flow}}$, where $\lambda$ is the same balancing weight.

\begin{table}[t]
\centering
\caption{\textbf{Performance on embodied reasoning benchmarks for spatial understanding across different model types.} The highest and second-highest accuracy values are highlighted in \textbf{bold} and \underline{underlined}, respectively. GEM-8B achieves state-of-the-art (SOTA) performance and near-SOTA competitive results across general-purpose and spatial specialist models.} \vspace{5pt}
\adjustbox{width=\linewidth}{
\begin{tabular}{lcccccc}
\toprule
Models~~~~~~~~~~~~~~ & \multicolumn{1}{c}{~~CV-Bench~~}
& \multicolumn{3}{c}{VSI-Bench} 
& \multicolumn{1}{c}{~~MMSI-Bench~~}
& \multicolumn{1}{c}{~~EmbSpatial~~}\\\cmidrule{3-5}
& All $\uparrow$
& ~Abs. Dist.~
& ~Rel. Dist.~
& ~All $\uparrow$~
& All $\uparrow$
& All $\uparrow$
\\
\midrule
% \multicolumn{7}{c}{\textit{Proprietary Models}} \\
% \midrule
Gemini-3-Pro~\citep{comanici2025gemini} & 82.5 & 42.8 & 56.6 & 53.0  & 45.9  & 81.0 \\
Seed1.8~\citep{seedseed1}  & 86.5 & 28.0 & 50.3 & 47.2 & 34.6  & 78.7 \\
GPT-4o~\citep{hurst2024gpt} & 78.6 & 5.3 & 37.0 & 34.0  & 30.3  & 71.9 \\
\midrule
% \multicolumn{7}{c}{\textit{Open-Source Vision-Language Models}} \\
% \midrule
Qwen3-VL-2B~\citep{bai2025qwen3} & 80.0 & 40.7 & 49.6 & 50.4 & 23.6 & 69.0 \\
Qwen3-VL-8B~\citep{bai2025qwen3} & 85.1 & 47.5 &  58.2 & 57.9  & 27.7 & 77.7\\
InternVL3.5-8B~\citep{wang2025internvl3} & 81.5 & 40.9 & 47.7 & 54.1 & 28.4  & 74.2 \\
LLava-OneVision-7B~\citep{li2024llava} & 61.9 & 20.2 & 42.5 & 32.4  & 26.6 & 73.1 \\
% BAGEL-MoT-7B~\citep{deng2025bagel} & 76.5 & 33.5 & 41.3 & 30.8 & 73.1 \\
\midrule
% \multicolumn{7}{c}{\textit{Spatial Specialist Models}} \\
% \midrule
SpaceR-7B~\citep{ouyang2025spacer} & 74.8  & 28.6 & 38.2 & 35.8 & 26.4  & 65.8 \\
VLM-3R~\citep{fan2025vlm}  & 71.8 & 49.4 & 65.4  & 60.7  & 27.9  & 68.2 \\
VST-7B~\citep{yang2025visual}& 83.5 & 43.8 & 60.0 &  60.6  & 32.6 & 73.6 \\
CambrainS-7B~\citep{yang2025cambrians} & 76.9 & 49.4 & 66.9 & 67.5 & 24.2  & 70.0 \\
SenseNova-SI-8B~\citep{sensenova-si} & 83.2 & 48.0 & 64.1 & 67.9  & \textbf{43.3} & 77.6 \\
\midrule
% \multicolumn{7}{c}{\textit{Ours}} \\
% \midrule
Qwen3-VL-2B-SFT & 80.7 & 45.1 & 62.2 & 60.0 & 28.9 & 72.7 \\
\rowcolor[HTML]{F1E6EC}
GEM-2B (Ours)  & 81.4 & 48.4 & 64.1 &  62.8  & 30.6 & 73.0 \\
Qwen3-VL-8B-SFT & \underline{85.6} & \underline{53.7} & \underline{71.4} &  \underline{68.6} & 32.8 & \underline{78.3}\\
\rowcolor[HTML]{F1E6EC}
GEM-8B (Ours) & \textbf{86.6} & \textbf{56.3} & \textbf{72.3} & \textbf{70.6} & \underline{35.3} & \textbf{79.4} \\
\bottomrule
\end{tabular}}
\label{tab:spatial_bench}
\end{table}

\section{Experiments}

\subsection{Implementation Details}
We adopt Qwen3-VL~\citep{bai2025qwen3} as our VLM backbone and Sana~\citep{xie2024sana} as the depth prediction head.  We define a light connector comprising 2 layers of MLP that bridge the backbone’s output space with the DiT’s input space. Since some of our training data lack ground-truth depth annotations, we use DepthAnythingv3~\citep{lin2025depth} to generate pseudo depth maps for supervision. We train for 500 steps in Stage 1, 4k steps in Stage 2, and 1 epoch in Stage 3. We set $\lambda=0.1$ to balance structural synthesis with semantic understanding. The training process is performed on 32 NVIDIA A800 GPUs, with a cosine learning rate scheduler from $1e-5$ to $1e-6$. In real-world VLA tasks, we adopt the dedicated action expert from RDT2~\citep{liu2026rdt2}.  For each specific task, we jointly fine-tune the entire framework for 50k steps on 8 NVIDIA A800 GPUs, with a linear scheduler and a learning rate of $1e-5$. The balancing weight $\lambda$ in VLA finetuning is also set to 0.1. More implementation details can be seen in the supplementary material.

\subsection{Evaluation on Embodied Reasoning Capacities}

\begin{table*}[t]
\centering
\caption{\textbf{Performance on the object placement and grounding spatial benchmarks across different model types.} The highest and second-highest accuracy values are highlighted in \textbf{bold} and \underline{underlined}. \textsuperscript{*} denotes results obtained from their reports. It shows that GEM-8B achieves the best performance, compared with general-purpose and embodied specialist models.} \vspace{5pt}
\adjustbox{width=\linewidth}{
\begin{tabular}{lccccccc}
\toprule
Models~~~~~~~~~~~~~~
& \multicolumn{3}{c}{~~~~~~~~RefSpatial~~~~~~~~} 
& \multicolumn{3}{c}{~~~~~~~~Where2Place~~~~~~~~}
& \multicolumn{1}{c}{RoboSpatial} \\
\cmidrule(lr){2-4}\cmidrule(lr){5-7}
& ~~~ Loc.~~~ & ~~~Pla.~~~ & ~~~All $\uparrow$~~~
& ~~~Seen ~~~ & ~~~Unseen ~~~& ~~~All $\uparrow$ ~~~
& All $\uparrow$ \\
\midrule
% \multicolumn{8}{c}{\textit{General-Purpose Models}} \\
% \midrule
Gemini-3-Pro~\citep{comanici2025gemini} & 30.0 & 35.0 & 34.3 & 58.6 & 43.3 & 54.0 & 57.4\\
Seed1.8~\citep{seedseed1} & 65.0 & 41.0 & 50.2 & 54.2 & 53.3 & 53.8 & 66.9 \\
GPT-4o~\citep{hurst2024gpt} & 8.00 &9.55 & 8.78 & 20.3 & 20.7 & 20.4 & 43.5 \\
\midrule
Qwen3-VL-2B~\citep{bai2025qwen3} & 36.0 & 29.0 & 27.4 & 45.7 & 43.3 & 45.0 & 40.7 \\
Qwen3-VL-8B~\citep{bai2025qwen3} & \underline{54.0} & 36.7 & 38.0 & 61.0 & 62.2 & 61.3 & 65.4 \\
% LLava-OneVision-7B~\citep{} & 61.9 & 32.4 & 27.5 & 26.6 & - \\
% InternVL3.5-8B & - & - & - & - & - & - & 45.7  \\

\midrule
% \multicolumn{8}{c}{\textit{Embodied Specialist Models}} \\
% \midrule

VeBrain-8B\citep{luo2025visual} & 0.03 &0.57 & 0.30 & 12.3 & 9.17 & 11.3 & 42.5 \\
Magma-8B~\citep{yang2025magma} & 1.00 & 8.00 &4.50 & 9.93 & 13.1 & 10.9 & 33.7\\
RoboBrain-2.0-7B\textsuperscript{*}~\citep{ji2025robobrain} & 36.0 & 29.0 & 32.5 & \underline{64.3} & 61.9 & \underline{63.6} &54.2\\
% Vlaser-8B & - & - & 59.2 & - & - & 69.5 & -\\
Mimo-Embodied-7B\textsuperscript{*}~\citep{hao2025mimo} & - & - & \textbf{48.0} & - & - & \underline{63.6} &  61.8 \\
Cosmos-Reason2-8B~\citep{azzolini2025cosmos} & 48.0 & 27.0 & 33.2 & 52.9 & 43.3 & 50.0 & \underline{65.7} \\
\midrule
% \multicolumn{8}{c}{\textit{Ours}} \\
% \midrule

Qwen3-VL-2B-SFT & 37.0 & 29.0 & 29.6 & 56.7 & 51.4 & 53.0 & 44.6 \\
\rowcolor[HTML]{F1E6EC}
GEM-2B (Ours) & 41.0 & 33.0 & 32.1 & 52.8 & 53.3 & 53.0 & 47.4 \\
Qwen3-VL-8B-SFT & 53.0 & \textbf{40.0} & \underline{45.8} & 60.0 & \underline{62.9} & 62.0 & 65.4 \\
\rowcolor[HTML]{F1E6EC}
GEM-8B (Ours) & \textbf{57.0} & \underline{38.0} & 44.4 & \textbf{65.7} &\textbf{63.3} & \textbf{65.0} & \textbf{66.9} \\

\bottomrule
\end{tabular}}\vspace{-10pt}
\label{tab:emboded_bench}
\end{table*}

We evaluate on public spatiotemporal embodied reasoning benchmarks, including CV-Bench~\citep{tong2024cambrian}, VSI-Bench~\citep{yang2024think}, MMSI-Bench~\citep{yang2025mmsi} and EmbSpatial~\citep{du2024embspatial}. As shown in Table \ref{tab:spatial_bench}, both scales of GEM yield significant improvements across all tasks compared to their base models, Qwen3-VL. For instance, on the challenging VSI-Bench and MMSI-Bench, GEM improves the scores by roughly 10\%. In particular, compared with open-source general-purpose baselines~\citep{bai2025qwen3,wang2025internvl3,li2024llava}, the 8B-scale variant of GEM achieves the strongest overall performance on the majority of benchmarks, and remains highly competitive on the remaining. Furthermore, it also achieves superior performance even when compared to spatial specialist models~\citep{fan2025vlm,ouyang2025spacer,yang2025visual,yang2025cambrians,sensenova-si}. These strong results highlight our model’s powerful spatiotemporal reasoning capabilities, which can be effectively transferred to downstream tasks.

We further evaluate our model on benchmarks, including RefSpatial~\citep{zhou2025roborefer}, Where2Place~\citep{yuan2024robopoint}, and RoboSpatial~\citep{song2025robospatial}, which focus on object placement and referring in embodied environments. The results summarized in Table~\ref{tab:emboded_bench} show that the 8B variant of GEM achieves the best overall performance compared with both general-purpose and embodied specialist models~\citep{luo2025visual,yang2025magma,ji2025robobrain,hao2025mimo,azzolini2025cosmos}, while the 2B variant also remains highly competitive with many larger-scale models~\citep{azzolini2025cosmos,yang2025magma,luo2025visual}. It is also worth noting that GEM exceeds the strong proprietary baseline, Gemini-3-Pro~\citep{team2025gemini}, by about 10\% on average across benchmarks. Such competence highlights GEM's outstanding spatial reasoning capabilities in embodied environments, making it a versatile backbone for embodied AI brains.

Moreover, to assess the impact of depth generative supervision, we exclude this component and fine-tune the base models on the constructed dataset, referred to as Qwen3VL-2B-SFT and Qwen3VL-8B-SFT. As shown in Table \ref{tab:spatial_bench} and Table \ref{tab:emboded_bench}, the performance of both scales drops compared to the full models. This is because the full model leverages generative supervision to integrate global structural features with semantic features within a shared representation space, which benefits stronger spatial perception capabilities. Notably, on distance-related questions in VSI-Bench~\citep{yang2024think}, the full models significantly outperform their counterparts, demonstrating that depth supervision enhances the model’s ability to perceive relative distance and spatial relationships.

\subsection{Evaluation on Downstream VLA tasks}

% In this section, we further examine how the enhanced capabilities of GEM transfer to improved performance when applied for downstream Vision-Language Action models (VLAs) in both simulation and real-world scenarios.

\begin{table}[tbp]
\centering
\caption{\textbf{Success rates on the LIBERO benchmark across four task suites.} It is demonstrated that GEM-VLA exhibits better performance than all baselines, suggesting that implicit geometry reasoning from depth generative supervision improves generalization across diverse manipulation tasks.} \vspace{5pt}
% \footnotesize
% \setlength{\tabcolsep}{10pt}
\adjustbox{width=\linewidth}{
\begin{tabular}{lccccc}
\toprule
Models~~~~~~~~~~~~~~~~~ & ~~~~~ Spatial~~~~~~ & ~~~~~ Object~~~~~~ & ~~~~~ Goal~~~~~  & ~~~~~ Long~~~~~  & ~~~~~ Average $\uparrow$~~~~~   \\
\midrule
% \multicolumn{6}{c}{\textit{Standard VLA Models}} \\
% \midrule
Diffusion Policy~\citep{chi2025diffusion} & 78.3 & 92.5 & 68.3 & 50.5 & 72.4 \\
Octo-Base~\citep{team2024octo}  & 78.9 & 85.7 & 84.6 & 51.1 & 75.1 \\
OpenVLA~\citep{kim2024openvla} &  84.7 & 88.4 & 79.2 & 53.7 & 76.5 \\
$\pi_{0}$ (reported)~\citep{intelligence2025pi_} & 96.8 & 98.8 & 95.8 & 85.2 & 94.2 \\
\midrule
% \multicolumn{6}{c}{\textit{Spatial-Enhanced VLA Models}} \\
% \midrule
TraceVLA~\citep{zheng2024tracevla} & 84.6 & 85.2 & 75.1 & 54.1 & 74.8 \\
SpatialVLA~\citep{qu2025spatialvla} & 88.2 & 89.9 & 78.6 & 55.5 & 78.1 \\
% CoT-VLA~\citep{zhao2025cot} &  81.5 & 90.6 & 76.6 & 67.6 & 83.9 \\
MolmoACT~\citep{lee2025molmoact} & 87.0 & 95.4 & 87.6 & 77.2 & 86.6 \\
DreamVLA~\citep{dreamvla25} &  97.5 & 94.0 & 89.5 & 59.5 & 92.6 \\
DepthVLA~\citep{yuan2025depthvla} & 96.4 & 98.0 & 95.8 & 89.2 & 94.9 \\
\midrule
Qwen3VL-SFT-VLA & 97.2 & 98.4 & 95.6 & 88.4 & 94.9 \\
\rowcolor[HTML]{F1E6EC}
GEM-VLA (Ours) & \textbf{99.0} & \textbf{98.8} & \textbf{97.1} & \textbf{89.3} & \textbf{96.1} \\
\bottomrule
\end{tabular}}
\vspace{-10pt}
\label{tab:libero_bench}
\end{table}

\subsubsection{Simulation Evaluation}

We perform evaluation on the widely adopted LIBERO benchmark~\citep{liu2023libero}. It comprises four task suites: Spatial, Object, Goal, and Long, each containing 10 diverse tasks with 50 trials per task. For comprehensive comparison, we select high-performance VLA models, including both standard models~\citep{chi2025diffusion,kim2024openvla,intelligence2025pi_,team2024octo} and spatially-enhanced VLA models~\citep{zheng2024tracevla,qu2025spatialvla,dreamvla25,lee2025molmoact,yuan2025depthvla} as baselines. Unlike prior models pre-trained on robot data, we fine-tune GEM-2B and its action expert from scratch for action prediction on LIBERO for 20k steps, following the StarVLA implementation~\citep{starvla2025}. The evaluation results, shown in Table \ref{tab:libero_bench}, demonstrate that GEM-VLA achieves the highest success rate across all task types. This indicates that despite large-scale action pretraining, standard VLAs still lack sufficient spatial grounding for precise manipulation. Furthermore, GEM-VLA also outperforms other spatially enhanced VLAs, showcasing the strongest physical grounding capacities. Additionally, the performance of the VLA based on GEM also surpasses that based on the standard SFT model, which further underscores the benefit of depth generative supervision for accurate action prediction.
\begin{figure}[tbp]
  \centering
  \includegraphics[width=\linewidth]{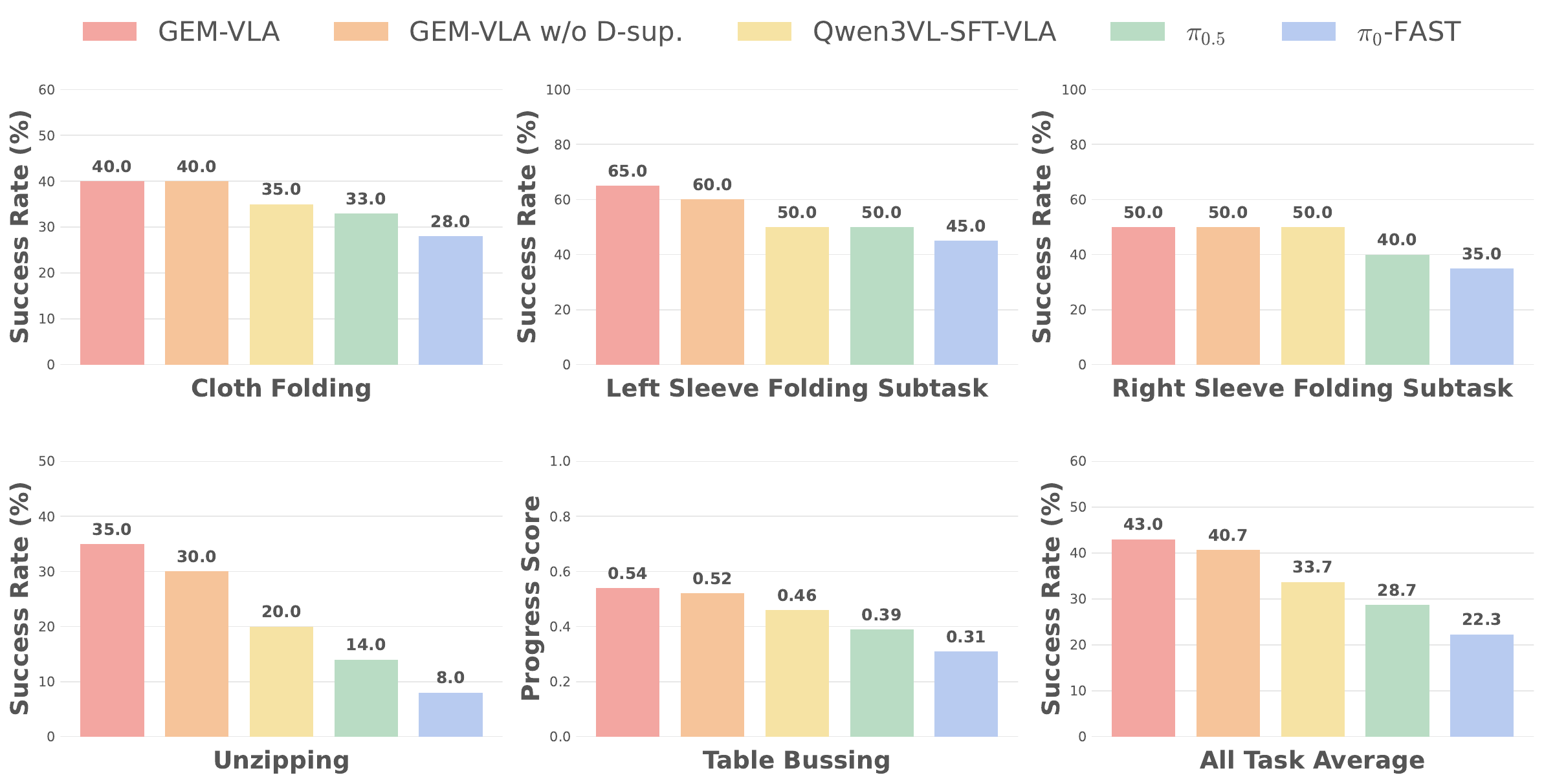} 
  \vspace{-15pt}
  \caption{\textbf{Comparison of GEM and baseline models on challenging real-world tasks.} The progress score
refers to the average percentage of sub-tasks completed in the long-horizon task. It is noted that the full GEM-VLA model outperforms previous baselines in success rate and progress score across all tasks and sub-tasks.}
  \label{fig:performance}
\end{figure}

\subsubsection{Real-World Evaluation}

\begin{figure}[htbp]
  \centering
  \includegraphics[width=\linewidth]{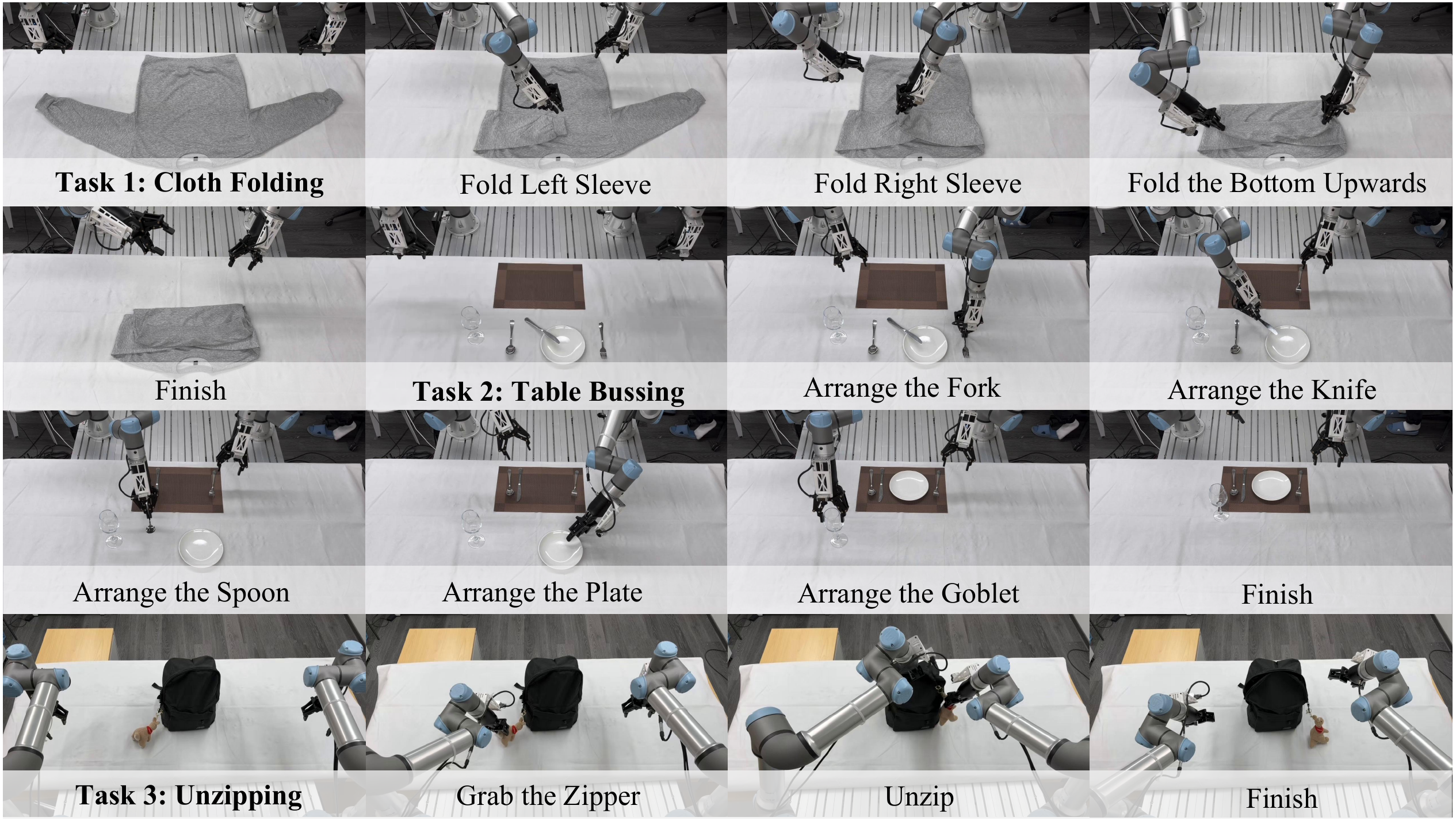}
  \caption{Demonstrations of three finetuned real-robot tasks, including one long-horizon task \textbf{table bussing} and two deformable object manipulation tasks \textbf{folding clothes} and \textbf{unzipping a backpack's zipper}. }
  \label{fig:task}
\end{figure}

We extend GEM to GEM-VLA following the approach in Sec.~\ref{sec:vla}, and deploy it on a UR5 platform to evaluate its performance on real-world manipulation tasks. We compare our model with most advanced baselines $\pi_0$-FAST~\citep{pertsch2025fast} and $\pi_{0.5}$~\citep{intelligence2025pi0.5_}. We finetune each model on several challenging real-world tasks, including long-horizon tasks (e.g., table bussing) and deformable object manipulation (e.g., folding clothes, unzipping a zipper). Task descriptions are shown in Figure \ref{fig:task}.

As summarized in Figure \ref{fig:performance}, our GEM-VLA demonstrates superior performance across all task and subtask categories. In both deformable manipulation tasks, GEM achieves higher success rates than all baselines~\citep{intelligence2025pi0.5_,pertsch2025fast}, particularly on the complex multi-step cloth folding task. For the long-horizon table bussing task, GEM-VLA achieves a significantly higher average progress score, which indicates its stronger long-horizon robustness and stability. These results confirm that GEM effectively transfers its pre-trained physical knowledge to state-of-the-art performance in diverse challenging real-world robotic manipulation. 

Moreover, we also investigate the effectiveness of depth generative supervision during VLA fine-tuning. We freeze the depth head and train only the VLM backbone and action expert with the action objective in Eq.~\ref{eq:action}. As shown in the Figure \ref{fig:performance}, performance drops on almost all tasks when depth generative supervision is removed, indicating that auxiliary depth prediction enables more accurate manipulation. Notably, the VLA fine-tuned from pre-trained GEM still outperforms its counterpart finetuned on the standard SFT model, which suggests that depth generative supervision facilitates learning physical priors in the embodied VLM, translating into stronger performance on downstream robot tasks.

\section{Ablation Studies}

\subsection{Superiority of Depth Supervision over RGB}

We investigate why depth is a more suitable supervisory signal compared to other alternatives. Specifically, we compare our depth-based generation with an RGB-based image generation task, where the model is trained to regenerate the input image.  For fair comparison, we fine-tune both models on the open-source VSI-590K dataset~\citep{yang2025cambrians}, keeping all other training hyper-parameters and strategies default. We evaluate each model on the representative spatial reasoning benchmarks, including CV-Bench~\citep{tong2024cambrian}, VSI-Bench~\citep{yang2024think}, and RoboSpatial~\citep{song2025robospatial}.

The results, summarized in Table \ref{tab:ablation} (rows 1 and 3), reveal that replacing depth supervision with RGB reconstruction leads to inferior performance, particularly on distance-related questions in VSI-Bench. This suggests depth provides more explicit cues about spatial relationships, such as relative distance, making it a more effective supervisory signal than RGB.

\subsection{Effectiveness of Progressive Training Strategy}

We evaluate the effectiveness of the proposed progressive three-stage training strategy. To compare, we perform a direct end-to-end training on both the depth generative head and understanding backbone using the VSI-590K dataset~\citep{yang2025cambrians}, and evaluate its performance on the same benchmarks~\citep{tong2024cambrian,yang2024think,song2025robospatial} above. The comparison results are shown in Table~\ref{tab:ablation}, in the second and third rows. It can be observed that direct end-to-end training underperforms the default three-stage training paradigm. This is because the generative head and connector fail to receive an appropriate initialization, which limits the effective fusion of semantic and structural features. Therefore, direct end-to-end co-training negatively affects the understanding model’s performance.

\begin{table}[tbp]
\centering
\caption{\textbf{Comparison between GEM and different settings.} The results show that the default model yields the best performance, which indicates depth supervision enhances structural learning compared to RGB and direct end-to-end training fails to effectively integrate semantic and structural features.} \vspace{5pt}
% \footnotesize
\adjustbox{width=\linewidth}{
\begin{tabular}{lccccc}
\toprule
Models ~~~~~~~~~ & \multicolumn{1}{c}{~~CV-Bench~~}
& \multicolumn{3}{c}{VSI-Bench}
& \multicolumn{1}{c}{~~RoboSpatial~~} \\
\cmidrule(lr){3-5} 
& ~~~~All $\uparrow$~~~~
& ~~~~Abs.Dist.~~~~
& ~~~~Rel. Dist. ~~~~
& ~~~~All $\uparrow$~~~~
& All $\uparrow$ \\
\midrule
RGB Supervision & 80.9 & 47.5 & 62.8 & 60.0 & 44.6 \\
Direct End-to-End Co-Training & 79.7 & 42.1 & 60.0 & 57.6 & 44.0 \\
\rowcolor[HTML]{F1E6EC}
Default Setting (GEM) & \textbf{81.1} & \textbf{47.8} & \textbf{65.2} & \textbf{63.0} & \textbf{48.9} \\
\bottomrule
\end{tabular}}
\label{tab:ablation}
\end{table}

\subsection{Effect of Generative Supervision on Structural Priors Learning}

\begin{figure}[htbp]
  \centering
  \includegraphics[width=\linewidth]{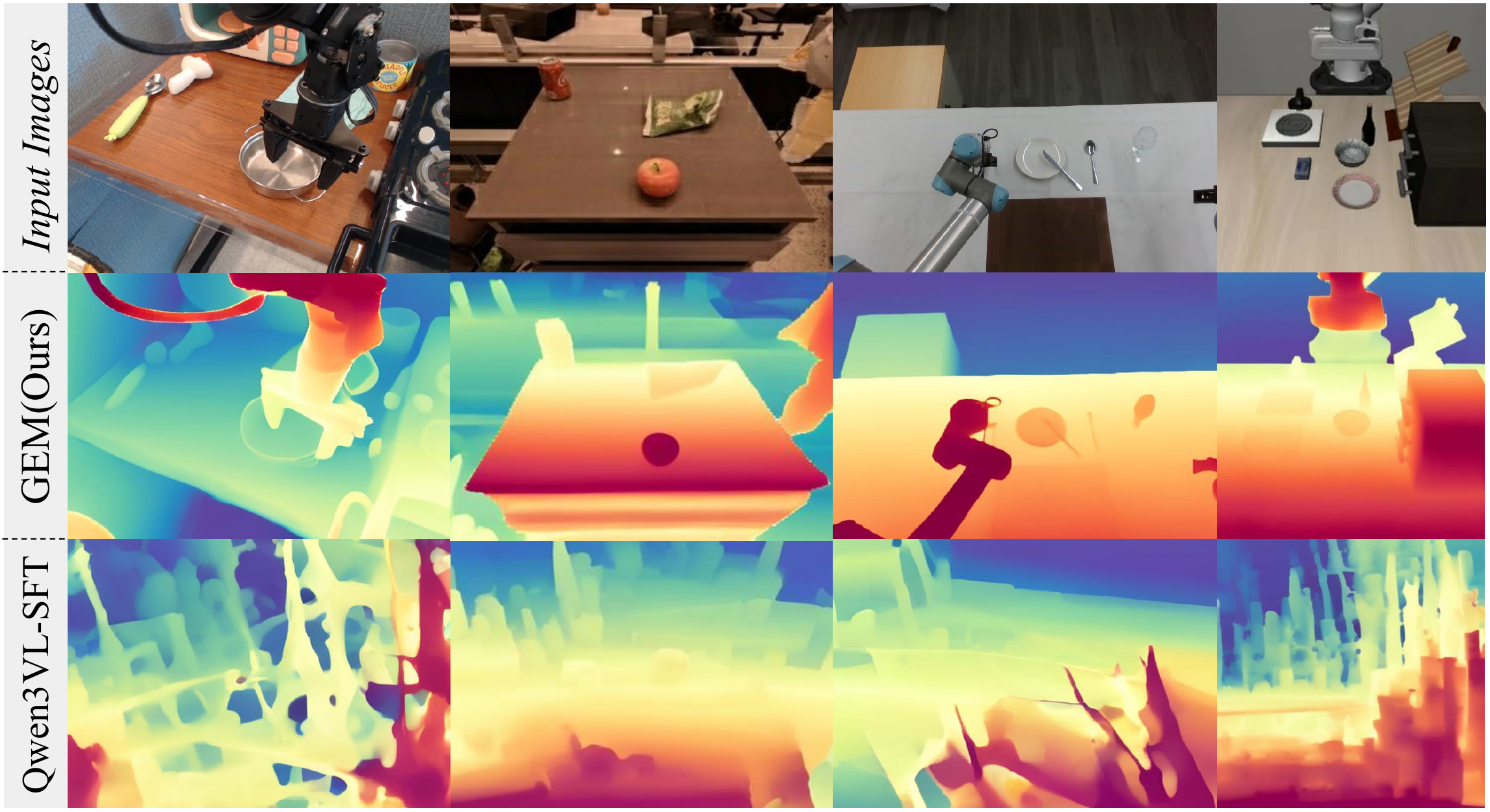}
  \caption{\textbf{Comparison of generation results from visual features in GEM and Qwen3-VL-SFT.} The results generated using only semantic features from the standard SFT model exhibit limited structural details, while GEM features produce high-quality results. This is because generative supervision helps the model capture structural cues that are beneficial for spatial perception in embodied environments.}
  \label{fig:depth}
\end{figure}

To assess whether depth generative supervision improves structural awareness in GEM, we respectively feed the final-layer visual token features from both Qwen3-VL-SFT and GEM into depth generator. As illustrated in Figure \ref{fig:depth}, the generated results of Qwen3-VL-SFT exhibit limited structural details. This indicates that visual representations learned under standard SFT are dominated by high-level semantic signals, while explicit spatial and geometric information is limited. This also accounts for the suboptimal embodied reasoning performance in standard SFT models. In contrast, GEM presents high-fidelity depth generation results, highlighting the effectiveness of depth generative supervision in capturing crucial low-level structural information from 2D inputs.

\section{Conclusion}

In this paper, we introduce GEM, a novel Generative-supervised Embodied vision-language framework designed to bridge the gap between high-level semantic reasoning and low-level physical grounding by learning depth generation as an intrinsic target for scene geometry.  A progressive training recipe is adopted to optimize depth synthesis and language objectives to better fuse structural and semantic representations. Beyond architectural and training advancements, we also construct a large-scale dataset that covers various embodied tasks to support training. Extensive evaluations demonstrate that GEM achieves strong performance across a variety of embodied benchmarks. Furthermore, the VLA model built on GEM sets new records on simulation benchmarks and shows robust generalization in real-world robotic tasks.

\newpage

% --- 修改参考文献标题并强制靠左 ---
\renewcommand{\refname}{References}
\renewcommand{\bibname}{References}
\renewcommand{\bibsection}{\section*{\raggedright \Large References}}
\makeatother
% ----------------------------------

\bibliographystyle{plainnat}
\bibliography{main}

\newpage
\appendix
\section*{Appendix}

\section{SimplerEnv Evaluation on WidowX Robot Tasks}
We further evaluate our approach in the SimplerEnv simulation environment~\cite{li2024evaluating}, specifically under the WidowX robot setup. The Simpler WidowX benchmark comprises four task suites, including \textit{Put Carrot on Plate}, \textit{Put Eggplant in Basket}, \textit{Put Spoon on Towel} and \textit{Stack Blocks}, which are designed to assess robustness to visual variations and precise manipulation. Following the StarVLA implementation~\cite{starvla2025}, we fine-tune GEM and its action expert from scratch on BridgeDataV2~\cite{o2024open} for 50k steps. We report the final success rate for each task suite, which is evaluated over 100 trials with different random seeds. For a comprehensive comparison, we benchmark our method against both standard VLAs~\cite{chi2025diffusion,team2024octo,intelligence2025pi_,kim2024openvla} and recent VLAs equipped with spatial priors~\cite{liu2025towards,yang2025vlaser,qu2025spatialvla,zheng2024tracevla,huang2025thinkact}. As shown in Table~\ref{tab:simplerenv_widowx}, GEM-VLA achieves the highest average success rate overall, establishing a robust foundation for sim-to-real transfer. Notably, the GEM-based VLA outperforms the VLA built on the standard SFT model, suggesting that GEM’s geometric priors help capture fine-grained spatial relationships and enable reliable action prediction in manipulation tasks.

\begin{table}[htbp]
\centering
\caption{\textbf{Success rates on the Simpler WidowX Robot benchmark.} The highest and second-highest accuracy values are highlighted in \textbf{bold} and \underline{underlined}, respectively. The results show that GEM-VLA achieves the highest average performance across all task types.}
\vspace{5pt}
\adjustbox{width=\linewidth}{
\begin{tabular}{lccccc}
\toprule
Model & ~~~Put Carrot~~~& ~~~Put Eggplant~~~ & ~~~Put Spoon~~~  &  ~~~Stack Block~~~ & ~~~Average~~~ \\
\midrule
Diffusion Policy~\cite{chi2025diffusion}     & 0\%  & 0\%    & 4.2\%    & 0\%    & 1.1\% \\
Octo-Base~\cite{team2024octo}      & 8.3\%  & 43.1\% & 12.5\% & 31.9\% & 16.0\% \\
OpenVLA~\cite{kim2024openvla}        & 0\%    & 4.1\%  & 0\%    & 0\%    & 1.0\% \\
$\pi_0$ ~\cite{intelligence2025pi_}    & \underline{55.8\%} & 79.2\% & 63.3\% & 21.3\% & 54.9\% \\
\midrule
RoboVLM~\cite{liu2025towards} & 25.0\% & 58.3\% & 29.2\% & 12.5\% & 31.3\% \\
TraceVLA~\cite{zheng2024tracevla} & 16.6\% & 65.0\% & 12.5\% & 16.6\% & 27.7\% \\
SpatialVLA~\cite{qu2025spatialvla} & 25.0\% & \textbf{100\%} & 16.7\% & \textbf{62.5\%} & 42.7\% \\
% UniVLA~\cite{bu2025univla} & 52.8\% & 55.6\% & 2.8\% & \textbf{80.6\%} & 47.9\% \\
ThinkAct~\cite{huang2025thinkact} & 37.5\% & 70.8\% & 58.3\% & 8.7\% & 43.8\%\\
% DepthVLA~\cite{yuan2025depthvla}  & 71.7\% & 89.2\%& 75.8\% & 62.5\%& 74.8\% \\
Vlaser~\cite{yang2025vlaser} & 52.5\% & \underline{87.9\%} & \underline{76.6\%} & 43.3\% & \underline{65.1\%} \\
\midrule
Qwen3VL-SFT-VLA & 44.0\% & 80.0\% & \textbf{82.0\%} & 40.0\% & 61.5\% \\
\rowcolor[HTML]{EBE9EE}
GEM-VLA (Ours) & \textbf{58.0\%} & 84.0\% & \textbf{82.0\%} & \underline{44.0\%} & \textbf{67.0\%} \\
\bottomrule
\end{tabular}}
\label{tab:simplerenv_widowx}
\end{table}

\section{More Implementation Details}
Built on Qwen3-VL~\cite{bai2025qwen3} as the VLM backbone, GEM integrates a lightweight 2-layer MLP connector and a DiT-based depth generator to combine generative supervision with representation learning seamlessly. GEM achieves embodied capabilities through a progressive three-phase training strategy: (i) initialize the connector, (ii) initialize the depth generator, (iii) end-to-end joint training. Detailed training configurations, including hyperparameters and optimization settings, are reported in Table~\ref{tab:vlm_train}. 

\begin{table}[t]

\caption{Detailed configuration for each training stage in VLM pretraining of GEM.}
\adjustbox{width=\linewidth}{
\begin{tabular}{lccc}
\toprule
 Configurations & ~~~~~~~~~~~~~~~~~~~Stage 1~~~~~~~~~~~~~~~~~~~ & ~~~~~~~~~~~~~~~~~~~Stage 2~~~~~~~~~~~~~~~~~~~ & ~~~~~~~~~~~~~~~~~~~Stage 3~~~~~~~~~~~~~~~~~~~  \\
\midrule
Batch Size & 128 & 128 & 128 \\
Learning Rate
& $1\times10^{-3}$ 
& $1\times10^{-4}$ 
& $1\times10^{-5}$ 
 \\
Training Steps & 500 steps & 4k steps & 1 epoch  \\
Optimizer & AdamW & AdamW &  AdamW \\
Weight Decay & 0.1 & 0.1 & 0.1  \\
Warmup Ratio & 0.00 & 0.00 & 0.03  \\
LR Schedule & Cosine & Cosine & Cosine  \\
Max Seq. Length & - & - & 16384 \\
GPU Nums & 32 &  32  & 32 \\
\bottomrule
\end{tabular}}
\label{tab:vlm_train}
\end{table}

\begin{figure}[htbp]
    \centering
    \begin{subfigure}[b]{0.48\textwidth}
        \centering
        \includegraphics[width=\textwidth]{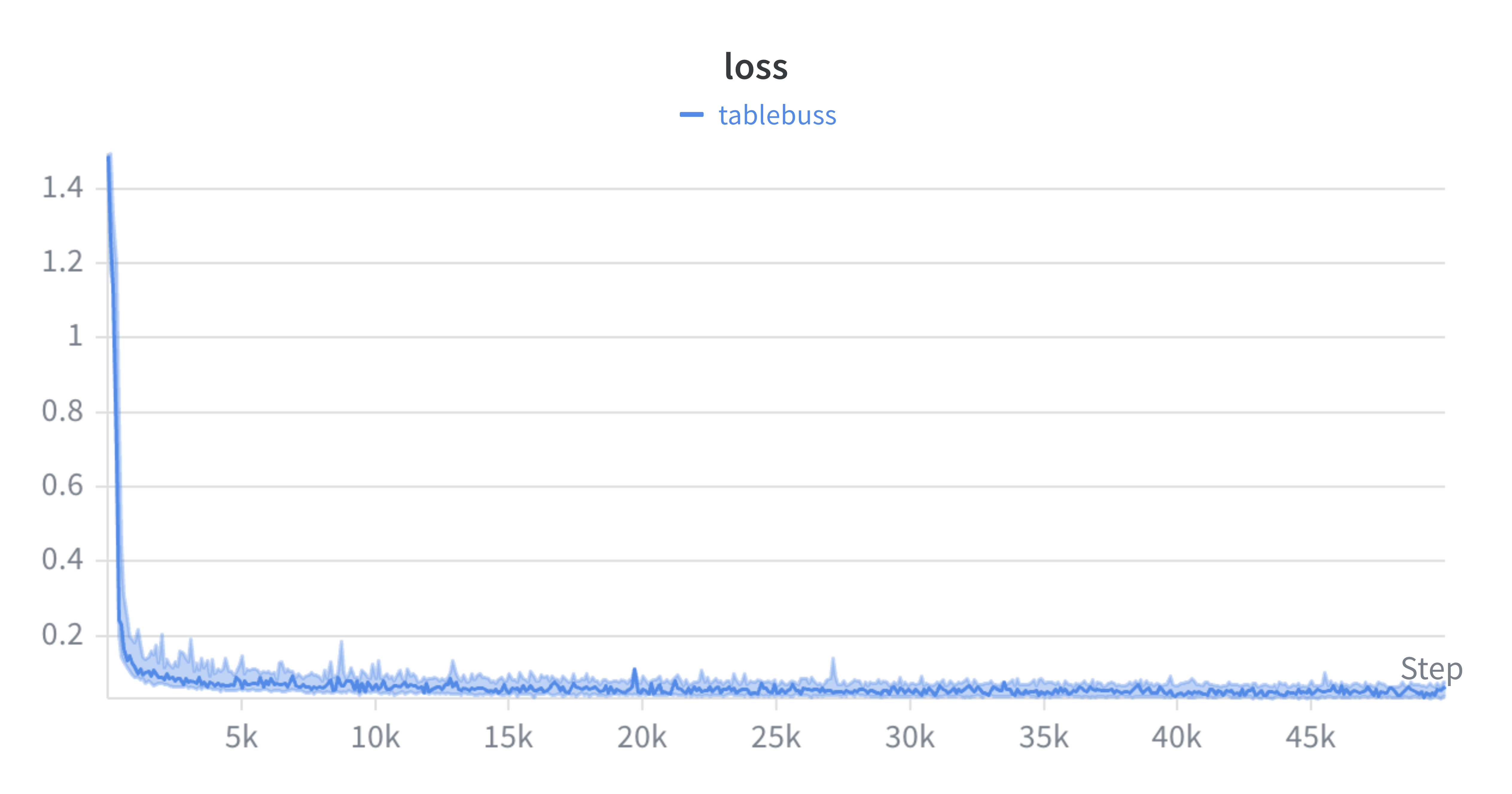}
        \caption{Loss curve on Table Bussing task.}
        \label{fig:table_buss_loss}
    \end{subfigure}
    \hfill
    \begin{subfigure}[b]{0.48\textwidth}
        \centering
        \includegraphics[width=\textwidth]{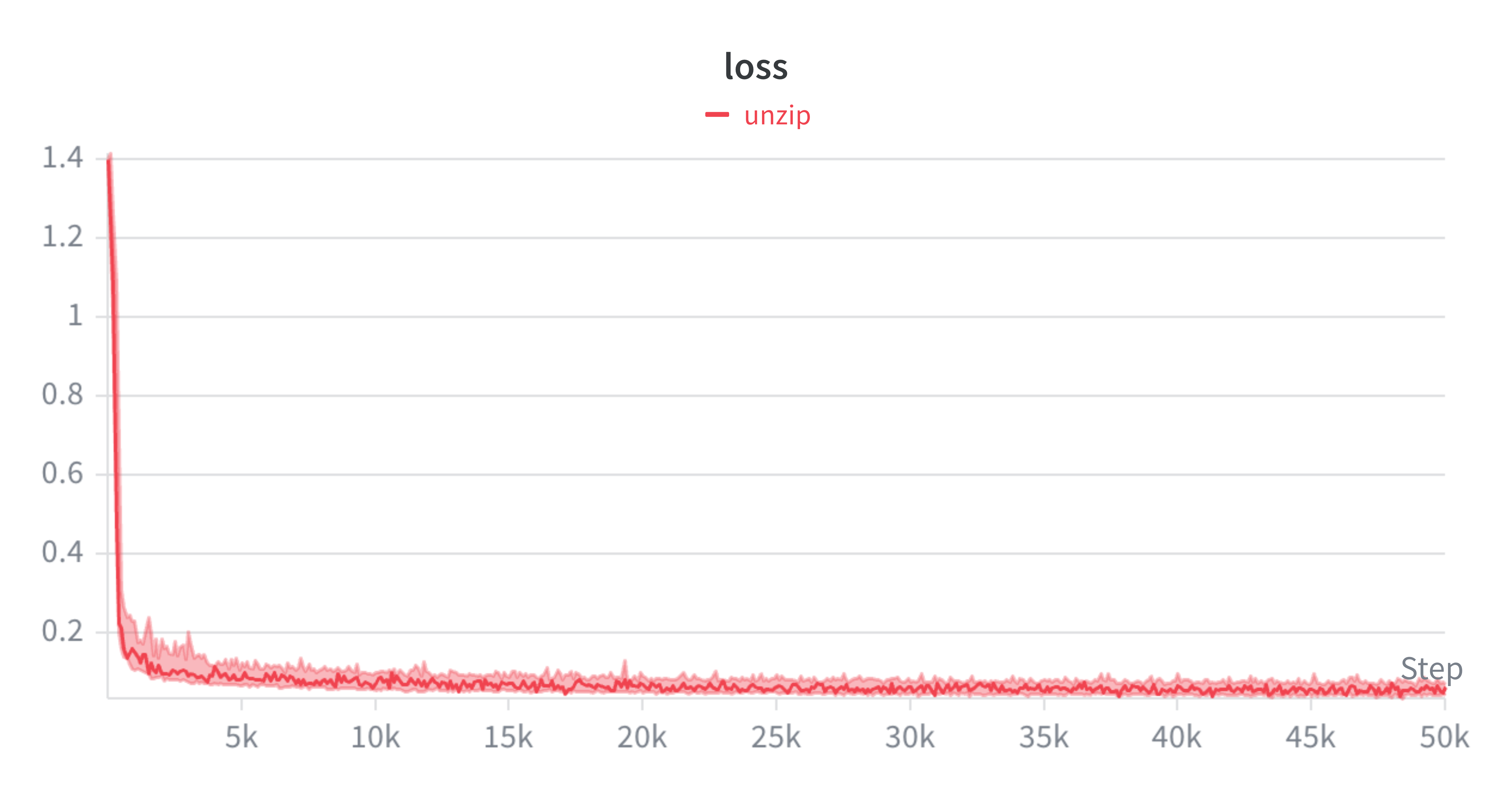}
        \caption{Loss curve on Unzipping task.}
        \label{fig:unzip_loss}
    \end{subfigure}
    \caption{Loss Curves of GEM-VLA on real-world task finetuning.}
    \label{fig:loss}
\end{figure}

To enable real-world robotic task evaluation, we extend GEM with a flow-based action expert built on the RDT2 implementation~\cite{liu2026rdt2}. For each specific task, we collect 200 trajectories and finetune the pretrained model for 50k steps with a global batch size of 256. We use an action chunk size of 32, and the observations consist of three camera views: one top view and two wrist views (left and right). The training loss curves are shown in Figure~\ref{fig:loss}. Performance is evaluated using both progress score and overall success rate, and results are averaged over 50 runs for each task.

For comparison, we re-implement two baseline models, $\pi_{0.5}$ and $\pi_0$-FAST, using the official OpenPI codebase. We train both baselines until convergence on 8 GPUs with a per-GPU batch size of 32. For $\pi_{0.5}$, we follow the official setup with discrete state inputs, an action horizon of 24, and a 32-dimensional action space. The training process uses bfloat16 precision and AdamW optimization. For $\pi_0$-FAST implementation, we keep the optimizer and scheduler settings the same as $\pi_{0.5}$.

\section{Details on GEM-4M construction}

\subsection{Embodied Grounding Data}
\begin{figure}[tb]
  \centering
  \includegraphics[width=\linewidth]{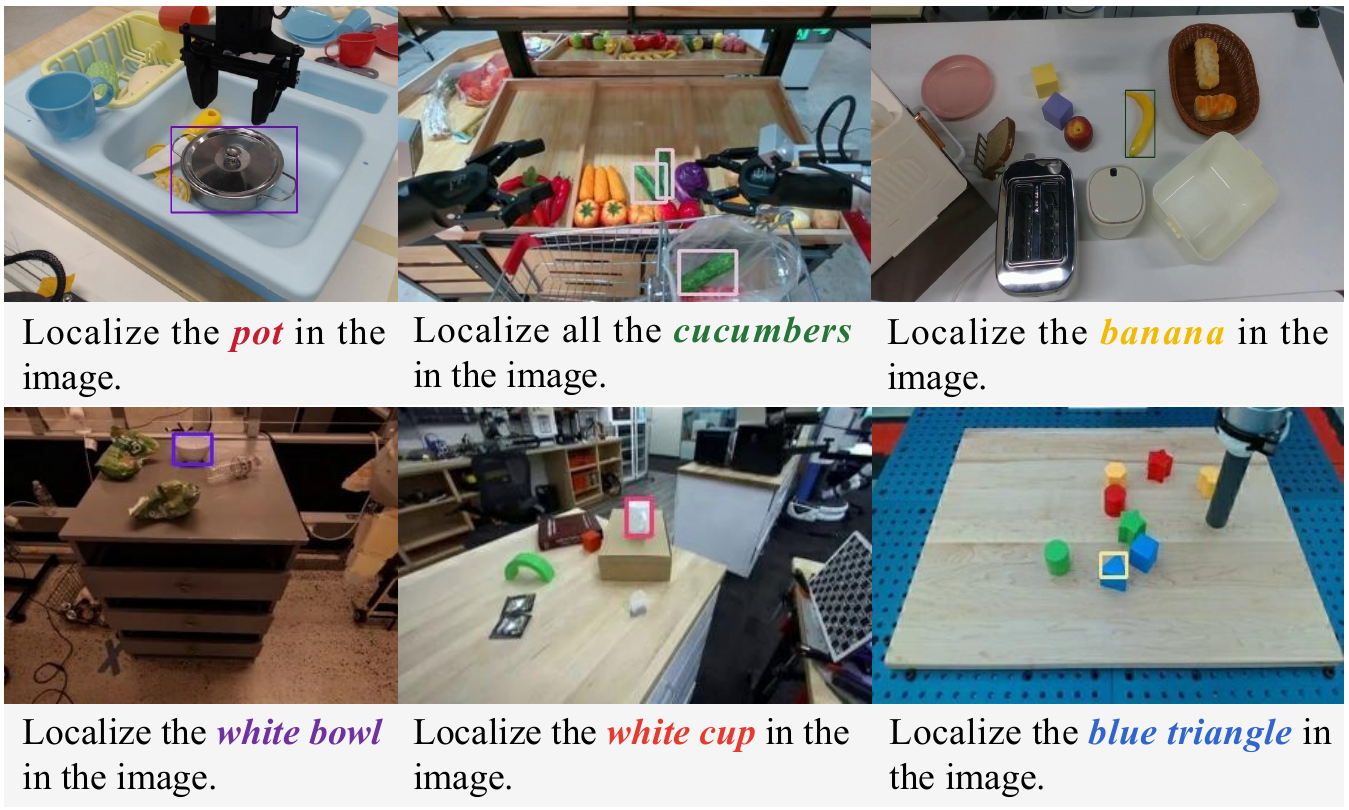}
  \caption{\textbf{Embodied Grounding Examples.} The target objects mentioned in the instructions are localized in the scene and highlighted with bounding boxes. }
  \label{fig:grounding}
\end{figure}

To further enhance grounding capabilities in physical manipulation scenarios, we generate an additional 100k high-quality data samples from open-source robot action datasets~\cite{wu2024robomind,o2024open,khazatsky2024droid}. The data generation process consists of two main stages. First, we extract the first frame of each robot operation video and employ Qwen3-VL~\cite{bai2025qwen3} to identify all object labels in its foreground, following the prompt templates provided in Figure \ref{fig:object_label_prompt}. Next, we use SAM3~\cite{carion2025sam3segmentconcepts} to obtain segmentation masks for each identified object label. To ensure annotation quality, only segmentation masks with confidence scores above 0.5 are retained. Since our annotations are in the form of bounding boxes or points, we derive bounding boxes by computing the minimum axis-aligned rectangle enclosing each mask, and obtain point annotations by randomly sampling a coordinate within the mask region. Some grounding examples are visualized in Figure \ref{fig:grounding}.

\subsection{Physical, Spatial Reasoning Data}

\begin{figure}[tb]
  \centering
  \includegraphics[width=\linewidth]{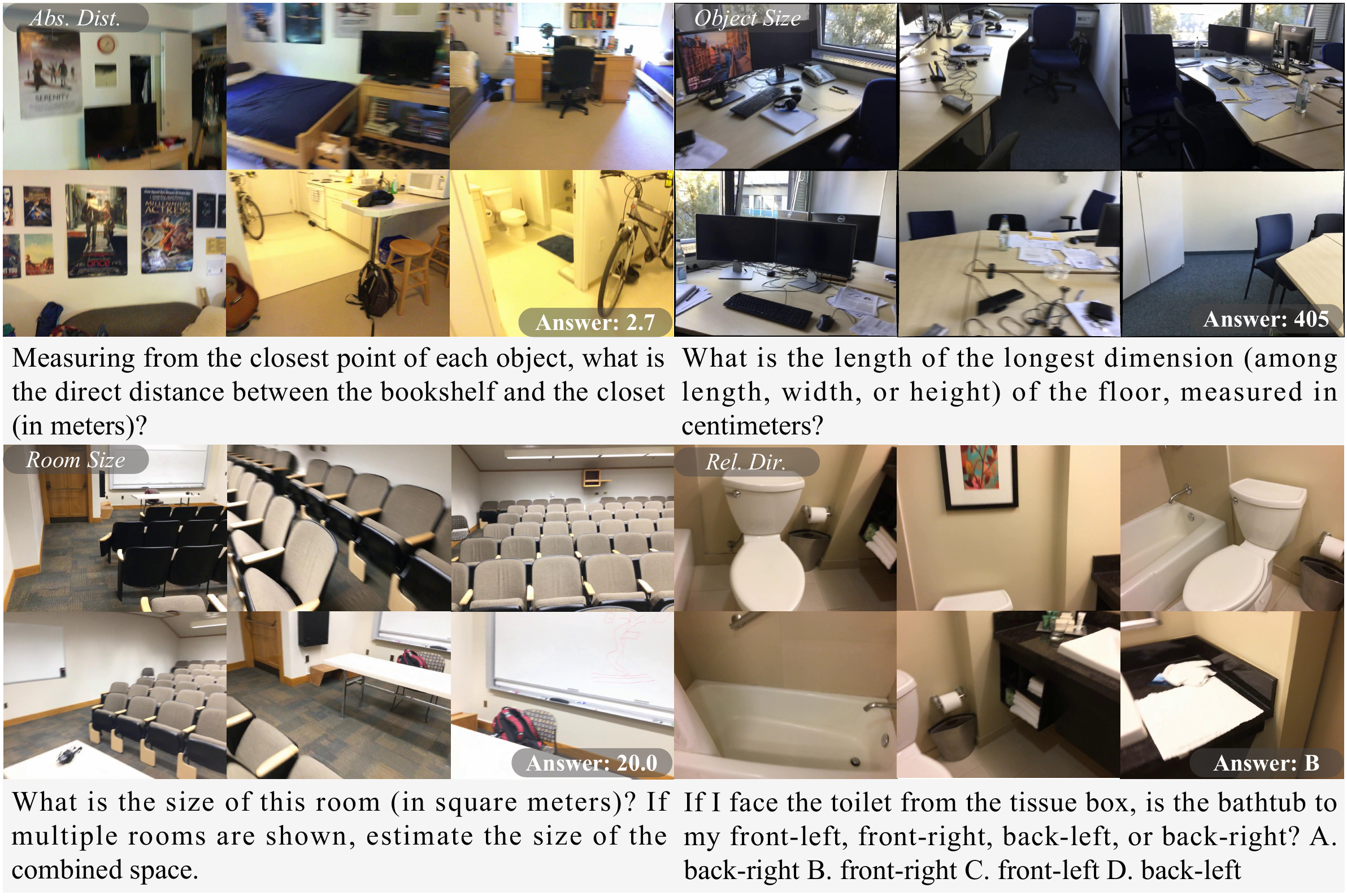}
  \caption{\textbf{Spatial Reasoning Examples.} Examples of generated spatial QA pairs on diverse reasoning tasks, including absolute distance estimation, object size prediction, room-size estimation, and relative direction understanding. }
  \label{fig:spatial}
\end{figure}

To enhance spatial intelligence capabilities, we manually construct a dataset of 100k 3D spatial perception samples derived from ScanNet~\cite{dai2017scannet}, Scannet++~\cite{yeshwanth2023scannet++} and ARKitScenes~\cite{baruch2021arkitscenes}, following methodologies established in VSI-Bench~\cite{yang2024think}. Specifically, we first convert each raw scene mesh to an Open3D~\cite{zhou2018open3d} point cloud and extract both spatial and semantic metadata from the associated annotations. These include room dimensions, center coordinates, counts of object categories, and 3D bounding boxes with rotation, extents, and centers for each object instance. Based on these representations, we generate QA pairs about layout properties and inter-object relationships, covering object counts, absolute and relative distances, object and room sizes, relative directions, and other spatial attributes, following the VSI-Bench question templates. Some generated examples are visualized in Figure \ref{fig:spatial}.

\subsection{Spatiotemporal Planning Data}
\begin{figure}[htbp]
  \centering
  \includegraphics[width=\linewidth]{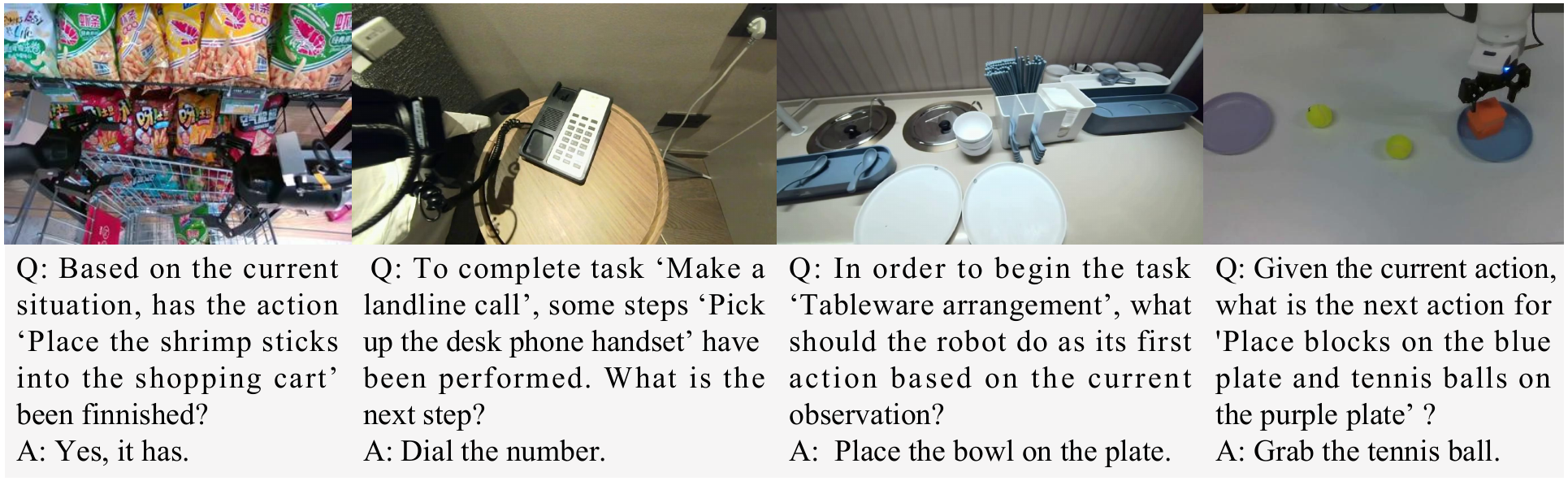}
  \caption{\textbf{Planning Examples.} Examples of generated planning QA pairs, including task completion verification, next step prediction conditioned on current observation, and initial step prediction for complex tasks. }
  \label{fig:planning}
\end{figure}

We collect robot videos from public datasets~\cite{wu2025robocoin,wu2024robomind,bu2025agibot_iros} with sub-task annotations and extract frames corresponding to each sub-task. Using these annotations, we generate question-answer pairs based on the RoboVQA~\cite{sermanet2024robovqa} template. Representative examples are shown in Figure~\ref{fig:planning}.

\begin{figure}[tbp]
  \centering
  \includegraphics[width=\linewidth]{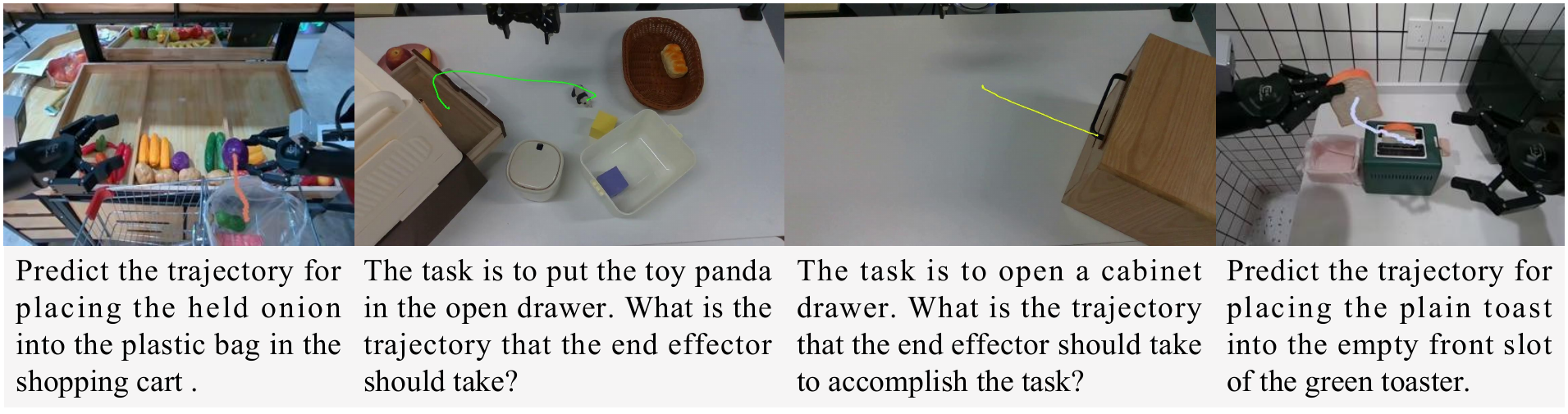}
  \caption{\textbf{Trajectory Examples.} The trajectories, composed of key trajectory points, represent the model-predicted paths for task completion. }
  \label{fig:trajectory}
\end{figure}

We also generate trajectory data to help the model learn object motion and action execution. For each sub-task video clip, we identify the manipulated object from the sub-task description using Qwen3~\cite{yang2025qwen3} with the prompt templates in Figure~\ref{fig:direct_object_prompt}. We then apply SAM3~\cite{carion2025sam3segmentconcepts} to the initial frame to obtain an instance mask, and use the object centroid as the initial state for trajectory tracking~\cite{karaev2025cotracker3}. The resulting trajectory is smoothed with cubic spline interpolation, and six uniformly spaced points are sampled as the final visual trace. Representative Examples are shown in Figure~\ref{fig:trajectory}.

\section{Limitation and Future Work}
Although GEM achieves strong performance on a wide range of embodied recognition benchmarks and robotic manipulation tasks, there remains room to further scale the model in terms of model size and training data. Moreover, the current GEM-VLA architecture has not been pretrained on large-scale robot datasets. As future work, we plan to incorporate large-scale robot data pretraining to equip the model with richer physical knowledge and further evaluate and refine the proposed methods.

\section{Simulation Rollouts Visualization}

\begin{figure}[tbp]
  \centering
  \includegraphics[width=0.9\linewidth]{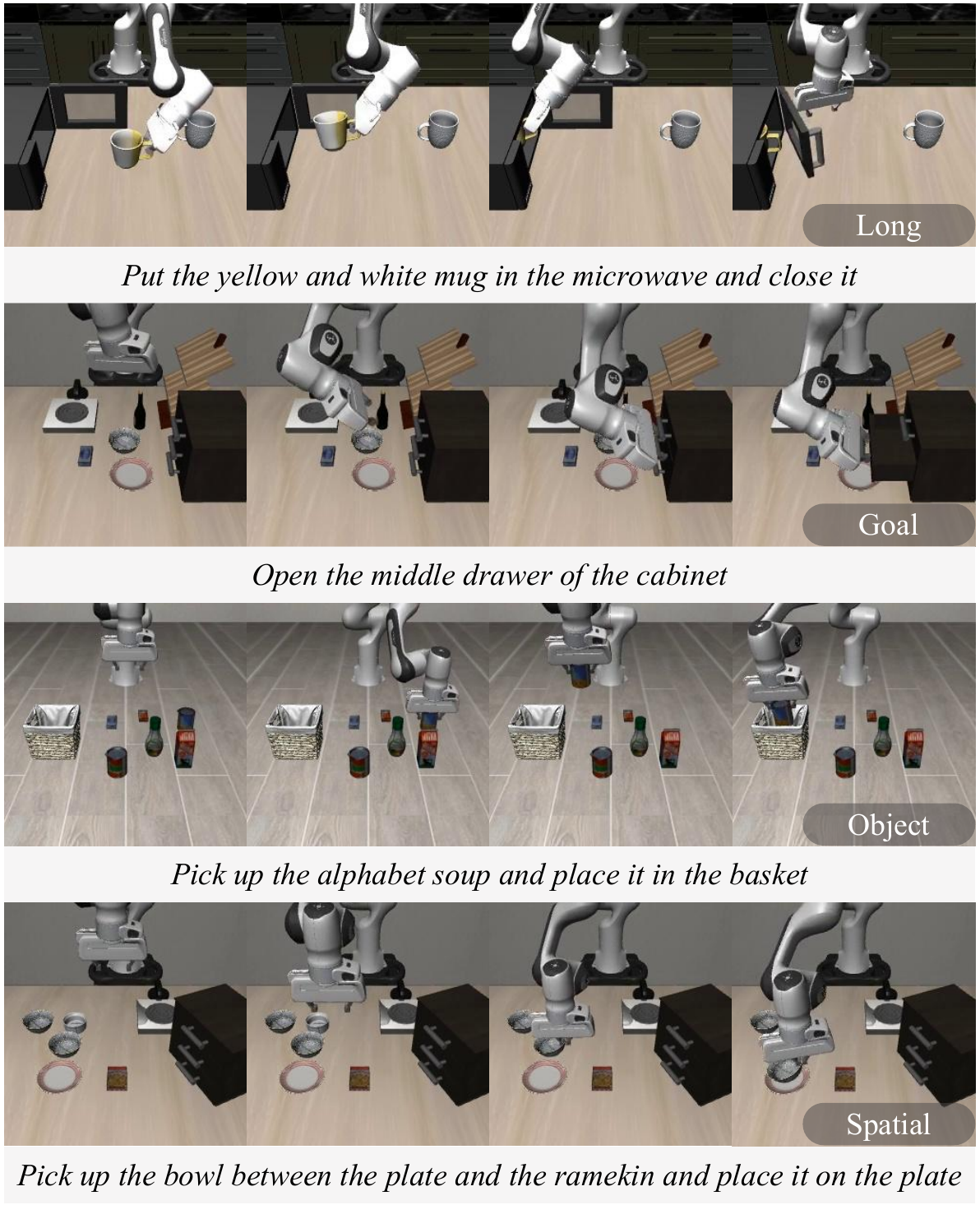}
  \caption{\textbf{Visualization of GEM-VLA's rollouts for LIBERO benchmark.}  }
  \label{fig:libero}
\end{figure}

\begin{figure}[tbp]
  \centering
  \includegraphics[width=\linewidth]{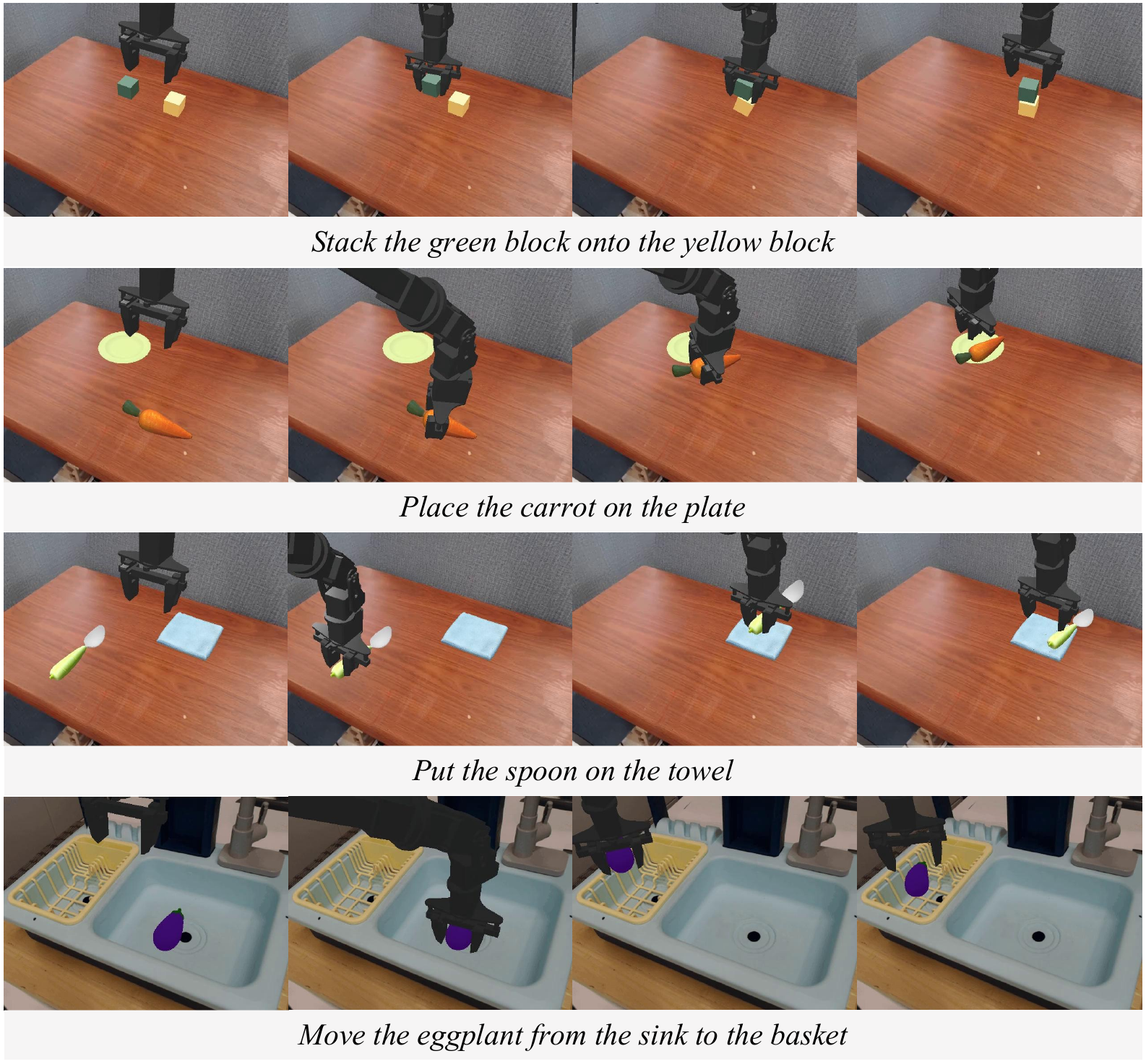}
  \caption{\textbf{Visualization of GEM-VLA's rollouts for Simpler WidowX benchmark.}  }
  \label{fig:simpler}
\end{figure}

In this section, we present qualitative visualizations of our model’s policy rollouts on simulation benchmarks. On LIBERO~\cite{liu2023libero}, we showcase successful rollouts on four representative manipulation tasks: LIBERO-Long, LIBERO-Goal, LIBERO-Object and LIBERO-Spatial in Figure \ref{fig:libero}. On SimplerEnv~\cite{li2024evaluating} with the WidowX robot setup, we further present rollouts on four task suites: Put Carrot on Plate, Put Eggplant in Basket, Put Spoon on Towel and Stack Blocks in Figure \ref{fig:simpler}. These qualitative results indicate that our model has strong potential for sim-to-real transfer.

\begin{figure}[htbp]
\centering
\begin{minipage}{0.98\linewidth}
\begin{tcolorbox}[
    colback=white,
    coltitle=white,
    fonttitle=\bfseries,
    title={Prompt for Object Labels Extraction},
    arc=2mm,
    boxrule=1.5pt,
    left=3mm, right=3mm, top=3mm, bottom=3mm
]

\textbf{Task}

First, identify and extract the objects mentioned in the instruction sentence. Then, based on the image, identify and list any additional objects that are clearly visible in the foreground or prominent areas of the image (such as on the table), excluding those already mentioned in the instruction sentence.

\vspace{0.5em}
\textbf{Extraction Rules}

\begin{enumerate}[leftmargin=*, noitemsep, topsep=2pt]
    \item \textbf{Scope}: Extract the full noun phrase representing the object, including all descriptive modifiers (adjectives, colors, sizes) and quantifiers.
    \item \textbf{Multiple Objects}: If there are multiple objects, list them in a comma-separated format.
    \item \textbf{Foreground Focus}: Only include objects that are clearly visible in the foreground or main area of the image, such as on the table. Avoid including objects in the background or irrelevant items.
\end{enumerate}

\vspace{0.5em}
\textbf{Output Format}

Return the objects as a JSON object with two keys:
\begin{itemize}[leftmargin=*, noitemsep, topsep=2pt]
    \item \texttt{instruction\_mentioned\_objects}: A list of objects mentioned in the instruction sentence.
    \item \texttt{additional\_objects\_in\_image}: A list of objects visible in the image but not mentioned in the provided sentences.
\end{itemize}

\vspace{0.5em}
\textbf{Example} \\
\textbf{Input:} Pick up the blue ring from the table and put it in the wooden tray \\
\textbf{Output:} \\
\texttt{\{ \\
\hspace*{1em} "instruction\_mentioned\_objects": ["blue ring", "wooden tray"], \\
\hspace*{1em} "additional\_objects\_in\_image": ["red ring", "green ring"] \\
\}}

% \vspace{0.5em}

% \textbf{Example 2:} \\
% \textbf{Input:} Pick up the plastic spoon and scoop one bean from the bowl, pour the bean into the paper cup and put the plastic spoon back into the white bowl \\
% \textbf{Output:} \\
% \texttt{\{ \\
% \hspace*{1em} "instruction\_mentioned\_objects": ["plastic spoon", "bowl", "bean", "paper cup", "white bowl"], \\
% \hspace*{1em} "additional\_objects\_in\_image": ["coffee machine"] \\
% \}}

\end{tcolorbox}
\end{minipage}
\caption{Prompt template for object labels extraction.}
\label{fig:object_label_prompt}
\end{figure}
\newpage
\begin{figure}[htbp]
\centering
\begin{minipage}{0.98\linewidth}
\begin{tcolorbox}[
    colback=white,
    coltitle=white,
    fonttitle=\bfseries,
    title={Prompt for Direct Object Extraction},
    arc=2mm,
    boxrule=1.5pt,
    left=3mm, right=3mm, top=3mm, bottom=3mm
]

\textbf{Task}

Extract the \textbf{direct object} (the entity being acted upon, manipulated, or moved) from the given instruction sentence.

\vspace{0.5em}
\textbf{Extraction Rules}

\begin{enumerate}[leftmargin=*, noitemsep, topsep=2pt]
    \item Extract the full noun phrase representing the object, including all descriptive modifiers (adjectives, colors, sizes) and quantifiers.
    \item Do not include verbs, prepositions (e.g., in, on, at), or destination/location (e.g., in ``put apple on table'', extract only ``apple'').
    \item If no object is mentioned in the instruction, return ``robot arm''.
    \item If the instruction only mentions the action of the robot arm without specifying a manipulated object, return ``robot arm''.
    \item If the instruction mentions multiple objects being manipulated, return them separated by commas.
\end{enumerate}

\vspace{0.5em}
\textbf{Output Format}

Return only the extracted object text. Do not include punctuation, Markdown, or conversational filler.

\vspace{0.5em}
\textbf{Examples}

\textbf{Input:} Drag strainer forwards. \\
\textbf{Output:} strainer

\vspace{0.5em}

\textbf{Input:} Pick ketchup from table and place on oven. \\
\textbf{Output:} ketchup

\end{tcolorbox}
\end{minipage}
\caption{Prompt template for direct object extraction.}
\label{fig:direct_object_prompt}
\end{figure}

\end{document}